\documentclass{article}
\usepackage[T1]{fontenc}
\usepackage{arxiv}

\usepackage{amsmath,amsfonts,bm}

\def\eqref#1{equation~\ref{#1}}

\def\1{\bm{1}}

\DeclareMathAlphabet{\mathsfit}{\encodingdefault}{\sfdefault}{m}{sl}
\SetMathAlphabet{\mathsfit}{bold}{\encodingdefault}{\sfdefault}{bx}{n}

\usepackage{graphicx}
\usepackage{booktabs}
\usepackage{multirow}
\usepackage{xcolor}
\usepackage{natbib}
\setcitestyle{authoryear,round,citesep={;},aysep={,},yysep={;}}
\usepackage[hidelinks]{hyperref}
\hypersetup{
    pdftitle={Toward Unified Multimodal Representation Learning: Contrastive Alignment of Text, Image, and Point Cloud},
    pdfauthor={Ximeng Tao, Dimitar Filev, Gaurav Pandey}
}
\usepackage{url}

\title{\Large Toward Unified Multimodal Representation Learning:\\
Contrastive Alignment of Text, Image, and Point Cloud}
\renewcommand{\shorttitle}{{\fontsize{7}{8}\selectfont Toward Unified Multimodal Representation Learning: Contrastive Alignment of Text, Image, and Point Cloud}}
\date{}

\author{%
Ximeng Tao\textsuperscript{1}, Dimitar Filev\textsuperscript{1}, Gaurav Pandey\textsuperscript{2} \\[6pt]
\normalfont\textsuperscript{1}\,Department of Mechanical Engineering, Texas A\&M University \\
\normalfont\textsuperscript{2}\,Department of Engineering Technology and Industrial Distribution, Texas A\&M University \\
\normalfont College Station, TX~77843, USA \\[6pt]
\normalfont\small\texttt{ximeng@tamu.edu, dfilev@tamu.edu, gpandey@tamu.edu}}

\begin{document}
\raggedbottom

\maketitle

\begin{abstract}
Contrastive Language-Image Pre-training (CLIP) has shown impressive performance in aligning visual and textual representations. Recent works have extended this paradigm to 3D vision, where a common strategy is to align point-cloud representations with text and image features through pairwise contrastive objectives. However, such pairwise formulations optimize modality pairs separately rather than multimodal tuples jointly. In this paper, we propose Contrastive Tensor Pre-training (CTP), a joint contrastive learning framework that extends the pairwise similarity matrix to a multimodal similarity tensor for text--image--point-cloud alignment. We further study two key aspects of joint multimodal learning. First, we show that a simple geometric scorer can effectively align the three modality representations. Second, we identify partial-match negatives introduced by joint tuple construction and show that masking them improves joint alignment. Experiments on real-world automotive LiDAR from nuScenes and dense point clouds from ULIP-ShapeNet show that CTP outperforms a matched pairwise baseline in both zero-shot classification and instance-level retrieval. The learned representations also transfer to KITTI and Waymo without target-domain training, further supporting the cross-dataset generalization of CTP.
\end{abstract}

\section{Introduction}
\label{sec:intro}

\begin{figure}[!htbp]
    \centering
    \includegraphics[width=0.8\linewidth]{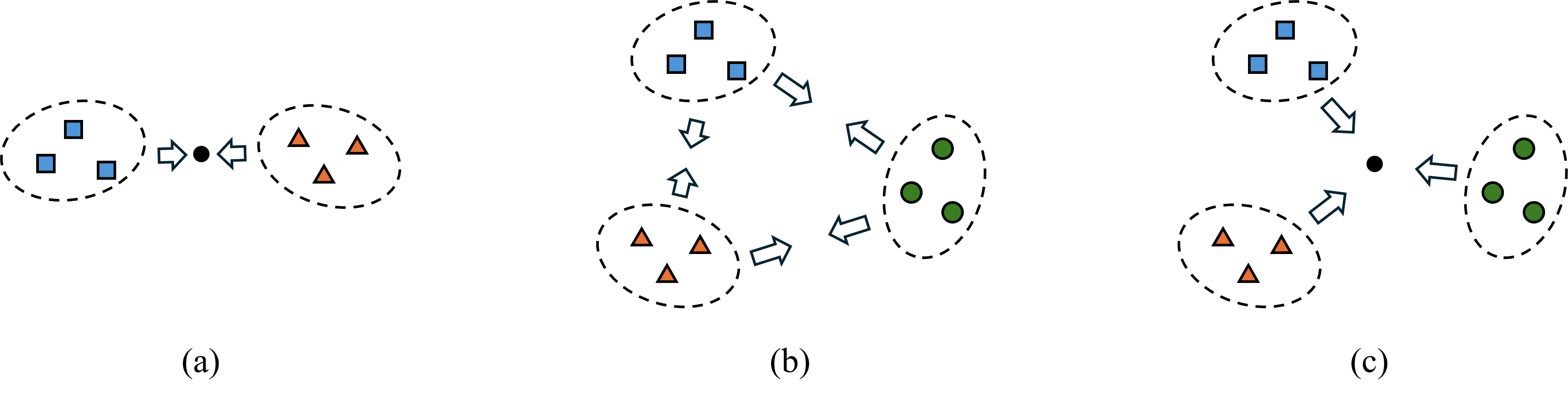}
    \caption{Overview of different contrastive representation learning methods.
    (a) CLIP: aligns two modalities.
    (b) Performs pairwise alignment between every modality pair.
    (c) CTP: aligns all modalities toward one point.}
    \label{Intro}
\end{figure}

Large Language Models (LLMs) have demonstrated remarkable performance in the textual modality, as well as strong capabilities in scene understanding, reasoning, and decision-making~\citep{brown2020language}.
Furthermore, Vision--Language Models (VLMs) have achieved the ability to understand visual and textual domains~\citep{liu2023visual, li2023blip, alayrac2022flamingo}, 
which significantly benefits many downstream tasks such as robotic manipulation~\citep{zitkovich2023rt, kim2024openvla} and end-to-end autonomous driving~\citep{chen2024driving, hwang2024emma,ma2025leapvad,xing2025openemma,zhou2025opendrivevla, huang2024drivegpt}.
Recently, there has been growing interest in extending the capabilities of LLMs from 2D to 3D domains~\citep{hong20233d, yang2025lidar, chen2024spatialvlm, kerr2023lerf}.
Point clouds are a common representation of 3D scenes, providing explicit geometric information and being less sensitive to illumination variations than camera images. These properties make point-cloud perception particularly important for applications such as autonomous driving.
However, real-world LiDAR point clouds are often sparse and incomplete~\citep{ma2024llms}, which makes their alignment with image and text representations particularly challenging.

CLIP~\citep{radford2021learning} provides an effective approach for aligning textual and visual modalities (Fig.~\ref{Intro}a). 
The pretrained Vision Transformer (ViT)~\citep{dosovitskiy2020image} can serve as the image encoder, forming the vision tower for LLMs~\citep{liu2023visual}.
This shared embedding approach has subsequently motivated extending vision--language representations to 3D modalities.
Inspired by the CLIP framework, several studies have explored extending the CLIP embedding space to point clouds~\citep{zhang2022pointclip}, showing improved scene understanding~\citep{chen2023clip2scene, peng2023openscene, liao2024vlm2scene}, localization~\citep{shubodh2024lip}, semantic segmentation~\citep{wang2023transferring} and object detection~\citep{lu2023open, pan2024clip}. 
A common strategy is to use contrastive learning to align 3D representations with CLIP text embeddings~\citep{zhang2022pointclip} or image embeddings~\citep{hess2024lidarclip}, or to align them with both modalities through similarity matrices~\citep{zeng2023clip2, hegde2023clip, xue2023ulip, wang2023large} (Fig.~\ref{Intro}b). 
Most existing language--image--3D methods optimize alignment through separate modality pairs. More recently, several methods have explored joint multimodal alignment by scoring complete multimodal tuples using statistical or geometric compatibility functions~\citep{saporta2024contrastingsymilesimplemodelagnostic, cicchetti2025triangleenablesmultimodalalignment, cicchetti2025gramianmultimodalrepresentationlearning}. These developments raise an important question for 3D multimodal representation learning: how should text, image, and point-cloud representations be jointly contrasted and scored within a shared embedding space?

To address this question, we propose Contrastive Tensor Pre-training (CTP), a joint contrastive learning framework for text, image, and point-cloud representations (Fig.~\ref{Intro}c). Instead of optimizing each modality pair independently, CTP organizes all text--image--point-cloud combinations in a similarity tensor and contrasts complete multimodal tuples within a unified candidate space. We further study two key aspects of this joint formulation. First, we investigate how multimodal tuples should be scored and show that a simple cosine-based geometric scorer remains effective in the three-modality setting. Although compressing the three pairwise relations into a scalar score can map different geometric configurations to the same value, the CTP scorer has a unique optimal angle configuration at multimodal co-location, where all three normalized representations coincide. This provides a simple and direct geometric constraint for three-modality alignment. Second, joint cross-entropy normalization introduces partial-match negatives along the anchor-associated rows and columns of the similarity tensor. Although these tuples are negative as a whole, they still contain a matched pair involving the anchor, so suppressing their scores can introduce competing alignment signals. Masking these candidates consistently improves the geometric joint methods considered in this work.

We evaluate CTP on real-world automotive LiDAR from nuScenes~\citep{nuscenes2019} and dense point clouds from ULIP-ShapeNet~\citep{xue2024ulip2scalablemultimodalpretraining}, and further assess cross-dataset transfer to KITTI~\citep{Geiger2012CVPR} and Waymo~\citep{Sun_2020_CVPR}. Our experiments show that CTP outperforms a matched pairwise baseline in both zero-shot classification and instance-level retrieval.
\section{Related Work}
\label{sec:related}

\paragraph{Multimodal Alignment.}
Multimodal alignment has gained increasing attention for its ability to enable models to handle diverse multimodal tasks. 
Vision--language pre-training~\citep{li2022blip, yao2021filip} has recently achieved remarkable success across various downstream applications. 
Pioneering works such as CLIP~\citep{radford2021learning} and ALIGN~\citep{jia2021scaling} focus on visual and textual modalities, 
demonstrating effective alignment between the two representations. 
Various methods have adapted CLIP to extract video~\citep{xu2021videoclip, luo2022clip4clip}, 
audio~\citep{guzhov2022audioclip, elizalde2023clap}, and 3D representations~\citep{zhu2023pointclip}. 
Beyond alignment between two modalities, ImageBind~\citep{girdhar2023imagebind} introduced a multimodal pre-training framework that leverages the image modality as a bridge to bind all modalities together. 
LanguageBind~\citep{zhu2023languagebind} instead employs the text modality as the alignment anchor to further enhance multimodal representation learning.
These approaches primarily establish multimodal alignment through a common anchor or separate modality pairs rather than directly scoring complete multimodal tuples.
More recently, joint multimodal learning has been explored by directly scoring complete multimodal tuples.
Symile~\citep{saporta2024contrastingsymilesimplemodelagnostic} uses a multilinear inner product to capture statistical dependence among multiple modalities, while TRIANGLE~\citep{cicchetti2025triangleenablesmultimodalalignment} and GRAM~\mbox{\citep{cicchetti2025gramianmultimodalrepresentationlearning}} introduce geometric similarity measures based on triangle area and Gram volume, respectively.
In contrast to these general multimodal formulations, we study joint contrastive learning specifically for text--image--point-cloud representations, with a focus on how multimodal tuples are scored and how contrastive candidates are constructed.

\paragraph{3D Representation Learning.}
Recently, there has been growing interest in connecting 3D point clouds with natural language through vision--language pre-training frameworks. 
Early works such as PointCLIP~\citep{zhang2022pointclip}, PointCLIP v2~\citep{zhu2023pointclip}, and CLIP2Point~\citep{huang2023clip2point} transfer knowledge from pretrained vision--language models to 3D representation learning.
LidarCLIP~\citep{hess2024lidarclip} extends this direction to real-world LiDAR by aligning LiDAR features with pretrained CLIP image representations. Subsequent methods, including CLIP$^2$~\citep{zeng2023clip2}, TriCLIP-3D~\citep{li2025triclip}, OpenShape~\citep{liu2023openshape}, and Uni3D~\citep{zhou2024uni3d}, further explore multimodal supervision for 3D representation learning. ULIP~\citep{xue2023ulip}, ULIP-2~\citep{xue2024ulip2scalablemultimodalpretraining}, and CLIP goes 3D~\citep{hegde2023clip} leverage both image and language supervision to align point-cloud representations with pretrained vision--language spaces.
JM3D~\citep{wang2023beyond} further combines image and language cues into a joint vision--language representation before aligning it with 3D features.
{
Beyond point clouds, multimodal learning has also been explored for meshes with MeshCLIP~\citep{song2023meshclip}, voxels with TriCoLo~\citep{ruan2024tricolo}, and 3D Gaussian Splatting with UniGS~\citep{li2025unigs} and TIGaussian~\citep{liu2026tigaussian}.
These methods extend vision--language alignment to diverse 3D representations.
Our work focuses on contrasting complete text--image--point-cloud tuples, investigating how tuple scoring and partial-match masking affect multimodal alignment.
}
\section{Methodology}
\label{sec:method}
\begin{figure}[!htbp]
    \centering
    \includegraphics[width=1\linewidth]{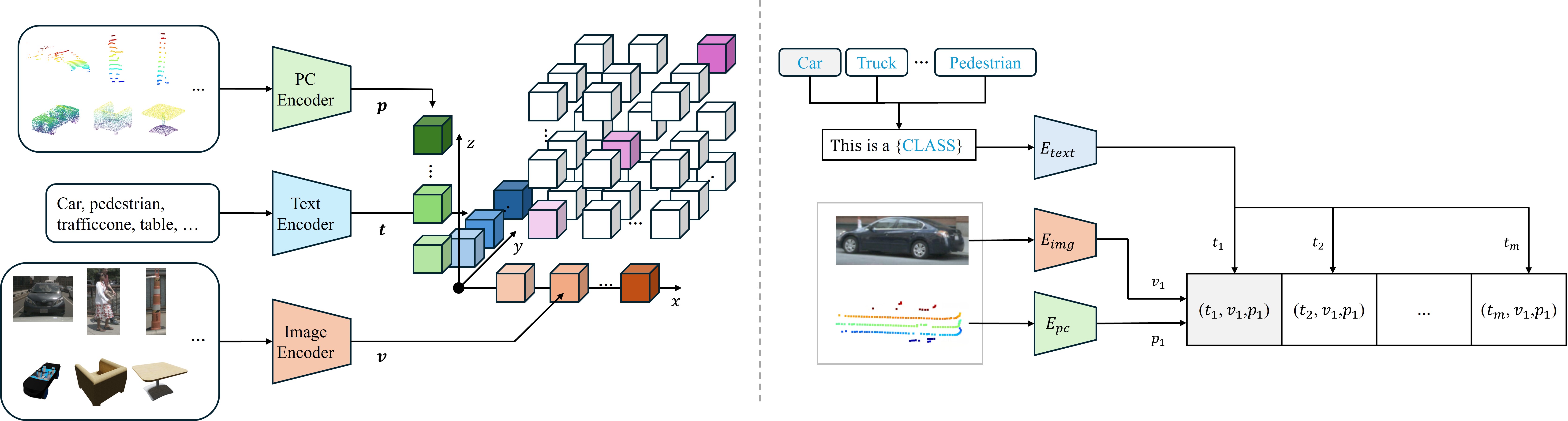}
    \caption{
    Overview of the CTP framework. Given matched text, image, and point-cloud inputs, modality-specific encoders map them into a shared embedding space. The resulting embeddings are arranged along three axes to construct a similarity tensor, where each entry represents the similarity of a text--image--point-cloud tuple. During zero-shot classification, class prompts are encoded as text candidates and jointly scored against the image and point-cloud representations to predict the class label.
    }
    \label{pipeline}
\end{figure}

In this section, we introduce Contrastive Tensor Pre-training (CTP), a similarity-tensor-based contrastive learning framework for jointly aligning text, image, and point-cloud representations. An overview of the framework is shown in Fig.~\ref{pipeline}. We first construct a joint multimodal candidate space using a similarity tensor, then define the tuple similarity scorer and tensor contrastive objective, and finally describe how partial-match negatives are handled.

\subsection{Similarity Tensor}
\label{S_tensor}
Contrastive alignment between two modalities is commonly formulated using a pairwise similarity matrix~\citep{radford2021learning}. For more than two modalities, a straightforward extension is to optimize separate pairwise contrastive objectives for each modality pair~\citep{xue2023ulip,zeng2023clip2,hegde2023clip}. For $q$ modalities and a mini-batch of size $b$, this formulation produces $\frac{q(q-1)}{2}$ similarity matrices, each containing $b^2$ pairwise scores. In contrast, the Cartesian product of the same mini-batch contains $b^q$ possible $q$-way tuples, each jointly indexed by one sample from every modality. Pairwise and joint formulations therefore differ in the structure of their contrastive candidate spaces: pairwise objectives normalize candidates within individual modality pairs, whereas joint objectives normalize complete multimodal tuples as contrastive candidates~\citep{saporta2024contrastingsymilesimplemodelagnostic}.

We focus on three modalities in this work: text, image, and point cloud, denoted by $T$, $I$, and $P$, respectively. Given a mini-batch of matched triplets,
\begin{equation}
\mathcal{B}=\left\{(T_i,I_i,P_i)\right\}_{i=1}^{b},
\end{equation}
the three modality-specific encoders produce unit-normalized representations
\begin{equation}
\mathbf{t}_i
=
\frac{E_{\mathrm{text}}(T_i)}
{\left\lVert E_{\mathrm{text}}(T_i)\right\rVert_2},
\qquad
\mathbf{v}_i
=
\frac{E_{\mathrm{img}}(I_i)}
{\left\lVert E_{\mathrm{img}}(I_i)\right\rVert_2},
\qquad
\mathbf{p}_i
=
\frac{E_{\mathrm{pc}}(P_i)}
{\left\lVert E_{\mathrm{pc}}(P_i)\right\rVert_2},
\label{eq:normalized_features}
\end{equation}
where $\mathbf{t}_i,\mathbf{v}_i,\mathbf{p}_i\in\mathbb{R}^{d}$, with $d$ denoting the dimensionality of the shared embedding space.

\paragraph{Pairwise contrastive learning.}
The conventional pairwise formulation optimizes three independent bidirectional contrastive objectives:
\begin{equation}
\mathcal{L}_{\mathrm{pair}}
=
\frac{1}{3}
\left(
\mathcal{L}^{\mathrm{pair}}_{T\leftrightarrow I}
+
\mathcal{L}^{\mathrm{pair}}_{T\leftrightarrow P}
+
\mathcal{L}^{\mathrm{pair}}_{I\leftrightarrow P}
\right).
\label{eq:pair_loss}
\end{equation}
For example, the text--image objective is
\begin{equation}
\mathcal{L}^{\mathrm{pair}}_{T\leftrightarrow I}
=
\frac{1}{2}
\left(
\mathcal{L}^{\mathrm{pair}}_{T\rightarrow I}
+
\mathcal{L}^{\mathrm{pair}}_{I\rightarrow T}
\right),
\label{eq:pair_ti_bidirectional}
\end{equation}
where the text-to-image term follows the standard CLIP objective~\citep{radford2021learning}:
\begin{equation}
\mathcal{L}^{\mathrm{pair}}_{T\rightarrow I}
=
-\frac{1}{b}
\sum_{i=1}^{b}
\log
\frac{
\exp\left(\mathbf{t}_i^\top\mathbf{v}_i/\tau\right)
}{
\sum_{j=1}^{b}
\exp\left(\mathbf{t}_i^\top\mathbf{v}_j/\tau\right)
}.
\label{eq:pair_t2i}
\end{equation}
Here, $\tau$ denotes the temperature parameter. For a fixed text anchor $\mathbf{t}_i$, Eq.~\ref{eq:pair_t2i} contrasts the matched image $\mathbf{v}_i$ against $b-1$ mismatched image candidates, yielding $b$ candidates in total. The reverse direction and the other modality pairs are defined analogously. We use this formulation as the matched pairwise baseline in our experiments.

\paragraph{Tensor contrastive learning.}
CTP instead constructs a joint candidate space over complete text--image--point-cloud tuples. We define a similarity tensor
\begin{equation}
\mathbf{S}\in\mathbb{R}^{b\times b\times b},
\qquad
S_{ijk}
=
s(\mathbf{t}_i,\mathbf{v}_j,\mathbf{p}_k),
\label{eq:similarity_tensor}
\end{equation}
where $s(\cdot,\cdot,\cdot)$ denotes a tuple similarity scorer. The indices $i$, $j$, and $k$ independently index the text, image, and point-cloud samples in the mini-batch, respectively. Each entry $S_{ijk}$ therefore represents the similarity of one text--image--point-cloud tuple. The diagonal entries $S_{iii}$ correspond to matched triplets, while the remaining entries form tuple-level negative candidates with different degrees of cross-modal matching.

Analogous to pairwise contrastive learning, we define a directional tensor loss by fixing one modality sample as the anchor. Fixing the text sample $\mathbf{t}_i$ leaves the image and point-cloud indices free, forming a two-dimensional candidate plane $\mathbf{S}_{i,:,:}\in\mathbb{R}^{b\times b}$. The text-anchored tensor loss over the complete candidate plane is
\begin{equation}
\mathcal{L}^{\mathrm{tensor}}_{T\rightarrow(I,P)}
=
-\frac{1}{b}
\sum_{i=1}^{b}
\log
\frac{
\exp\left(S_{iii}/\tau\right)
}{
\sum_{j=1}^{b}
\sum_{k=1}^{b}
\exp\left(S_{ijk}/\tau\right)
}.
\label{eq:tensor_t2ip}
\end{equation}
For each text anchor, Eq.~\ref{eq:tensor_t2ip} contrasts the matched tuple against all $b^2$ image--point-cloud combinations in the corresponding plane. In comparison, the pairwise text-to-image objective in Eq.~\ref{eq:pair_t2i} contains only $b$ image candidates for each text anchor.

The other two anchor directions are defined analogously by fixing the image or point-cloud sample. The complete tensor contrastive objective averages the three directional losses:
\begin{equation}
\mathcal{L}_{\mathrm{tensor}}
=
\frac{1}{3}
\left(
\mathcal{L}^{\mathrm{tensor}}_{T\rightarrow(I,P)}
+
\mathcal{L}^{\mathrm{tensor}}_{I\rightarrow(T,P)}
+
\mathcal{L}^{\mathrm{tensor}}_{P\rightarrow(T,I)}
\right).
\label{eq:tensor_loss}
\end{equation}

\begin{figure}[!htbp]
    \centering
    \includegraphics[width=0.65\linewidth]{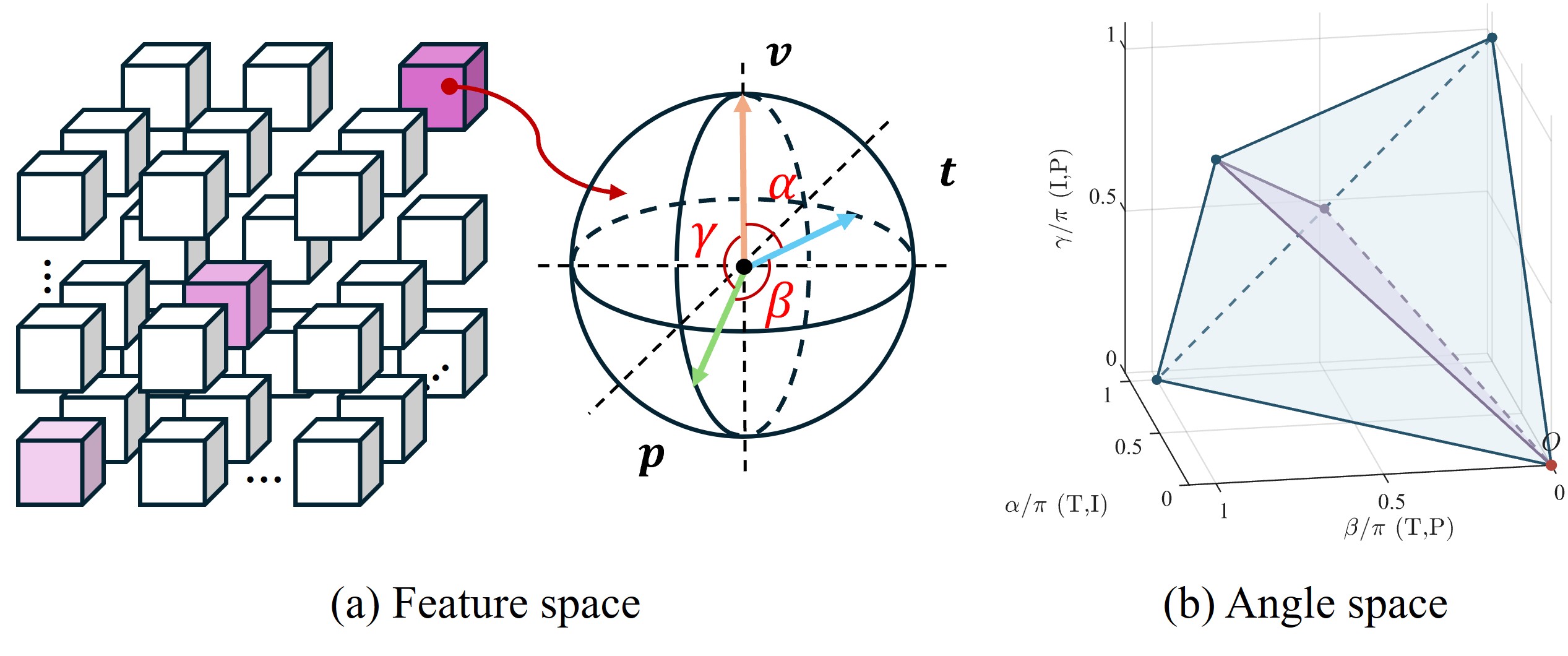}
    \caption{Geometric interpretation of the triplet angle space. (a) Feature space: three modality embeddings in a similarity-tensor tuple define three pairwise angles on the unit hypersphere. (b) Angle space: the three angles form a tetrahedral feasible region. The light purple plane corresponds to $\beta=\gamma=\theta$, and the point $O$ denotes $\alpha=\beta=\gamma=0$, where all three modality embeddings coincide.}
    \label{fig:angle_space}
\end{figure}

The key distinction between Eqs.~\ref{eq:pair_loss} and~\ref{eq:tensor_loss} lies in their contrastive candidate spaces. Fixing one modality in a pairwise objective leaves one sample index to enumerate, resulting in $b$ candidates. In the tensor objective, fixing one modality leaves two sample indices to vary jointly, resulting in $b^2$ multimodal tuple candidates. Pairwise contrastive learning therefore normalizes candidates from one modality at a time, whereas tensor contrastive learning jointly normalizes combinations of the other two modalities.

Importantly, this distinction is independent of the specific tuple scorer $s(\cdot,\cdot,\cdot)$. The similarity tensor determines \emph{which multimodal combinations are contrasted}, whereas the scorer determines \emph{how each tuple is scored}. We discuss the design of the tuple scorer in the next section, followed by the treatment of partial-match negatives in the tensor candidate space.
\subsection{Similarity Scorer}
\label{s_scorer}

Each entry of the similarity tensor corresponds to three unit-normalized modality representations, $\mathbf{t}_i$, $\mathbf{v}_j$, and $\mathbf{p}_k$. Their relative geometry can be characterized by three pairwise angles,
\begin{equation}
\alpha_{ij}
=
\arccos(\mathbf{t}_i^\top \mathbf{v}_j),
\qquad
\beta_{ik}
=
\arccos(\mathbf{t}_i^\top \mathbf{p}_k),
\qquad
\gamma_{jk}
=
\arccos(\mathbf{v}_j^\top \mathbf{p}_k).
\label{eq:triplet_angles}
\end{equation}
As shown in Fig.~\ref{fig:angle_space}, each multimodal tuple can therefore be represented by an angle configuration $(\alpha,\beta,\gamma)$ in a three-dimensional angle space. The geometrically feasible configurations form a tetrahedral domain.

Joint contrastive learning requires a scalar similarity score for each multimodal tuple. For a geometric scorer, this can be expressed as a mapping from the triplet angle configuration to a scalar,
\begin{equation}
S_{ijk}
=
f(\alpha_{ij},\beta_{ik},\gamma_{jk}).
\label{eq:angle_mapping}
\end{equation}
Since this mapping compresses a three-dimensional geometric configuration into a single scalar, different angle configurations may receive the same similarity score. The choice of $f$ therefore determines how tuple geometries are ranked and which configurations are favored during joint contrastive learning.

In CTP, we adopt a simple additive cosine scorer,
\begin{equation}
\begin{aligned}
S^{\mathrm{CTP}}_{ijk}&=
\mathbf{t}_i^\top \mathbf{v}_j
+
\mathbf{t}_i^\top \mathbf{p}_k
+
\mathbf{v}_j^\top \mathbf{p}_k \\
&=
\cos\alpha_{ij}
+
\cos\beta_{ik}
+
\cos\gamma_{jk}.
\end{aligned}
\label{eq:ctp_scorer}
\end{equation}
Substituting Eq.~\ref{eq:ctp_scorer} into Eq.~\ref{eq:tensor_t2ip}, the loss for a fixed text anchor $\mathbf{t}_i$ can be written as
\begin{equation}
\ell_i
=
\log
\sum_{j=1}^{b}
\sum_{k=1}^{b}
\exp
\left(
\frac{
S^{\mathrm{CTP}}_{ijk}
-
S^{\mathrm{CTP}}_{iii}
}{\tau}
\right).
\label{eq:ctp_anchor_loss}
\end{equation}
This loss encourages the matched text--image--point-cloud tuple to obtain a higher joint similarity than competing tuples in the same anchor plane. Although mapping the three-dimensional angle configuration $(\alpha,\beta,\gamma)$ to a scalar similarity score can assign the same value to different geometric configurations, the optimum of the CTP scorer is unique in the angle space. Since each pairwise cosine similarity is upper bounded by $1$, $S^{\mathrm{CTP}}_{ijk}$ reaches its maximum value $3$ only when $\alpha_{ij}=\beta_{ik}=\gamma_{jk}=0$, corresponding to the point $O$ in Fig.~\ref{fig:angle_space}, where $\mathbf{t}_i=\mathbf{v}_j=\mathbf{p}_k$. Thus, the cosine-sum scorer provides a simple geometric bias toward multimodal co-location despite the many-to-one mapping from angle configurations to scalar scores.

Alternative tuple scorers impose different notions of multimodal similarity. Symile~\citep{saporta2024contrastingsymilesimplemodelagnostic} uses a multilinear inner product to model joint statistical dependence and is not determined solely by the three pairwise angles. TRIANGLE~\citep{cicchetti2025triangleenablesmultimodalalignment} and GRAM~\citep{cicchetti2025gramianmultimodalrepresentationlearning} instead use geometric measures based on triangle area and Gram volume, respectively, which admit additional non-coincident optimum configurations. Detailed formulations and geometric comparisons of these scorers are provided in Appendix~\ref{app:scorer_analysis}.
\subsection{Tensor Contrastive Loss}
\label{s_tensor_loss}
\begin{figure}[!htbp]
    \centering
    \includegraphics[width=0.65\linewidth]{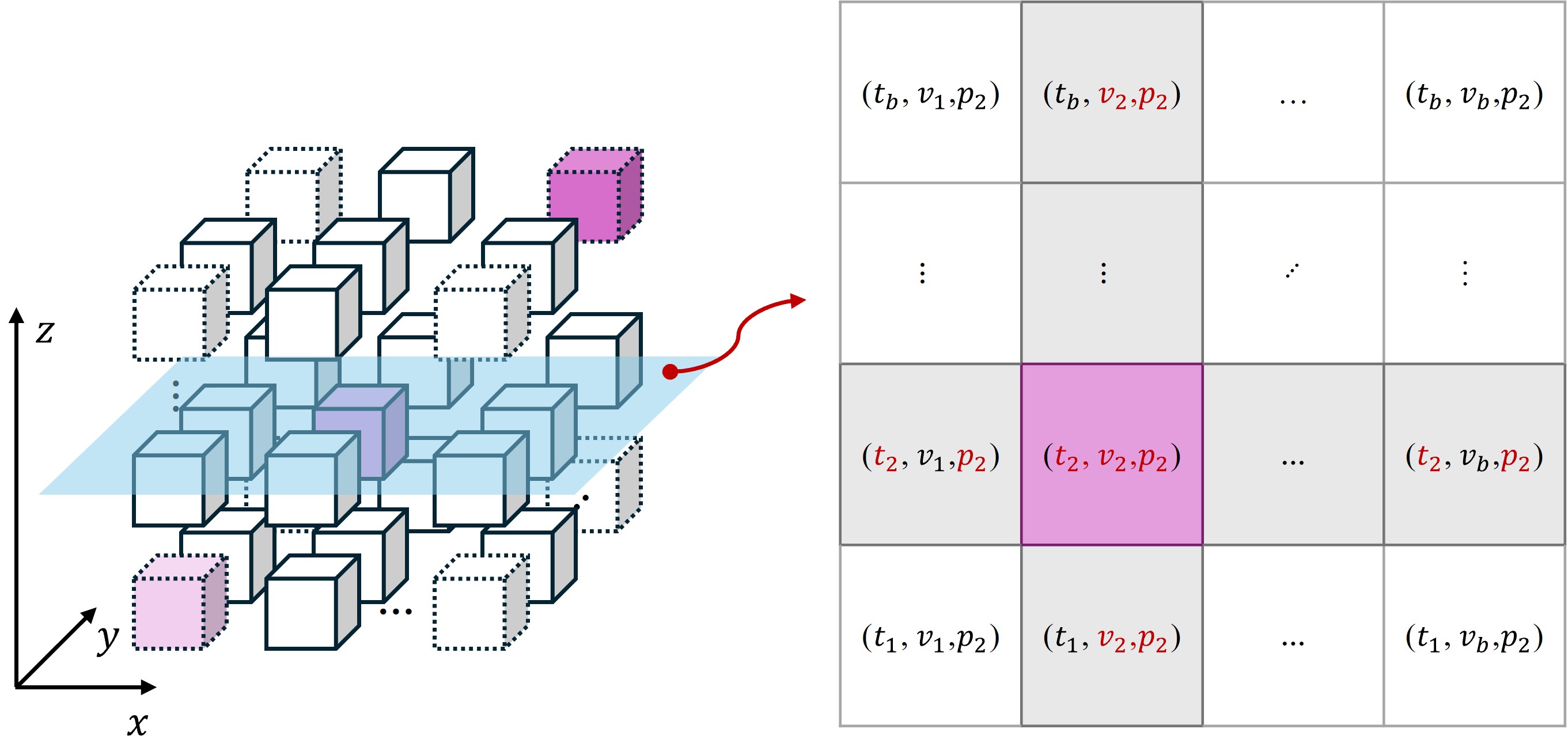}
    \caption{Masking partial-match negatives in an anchor plane. For a fixed point-cloud anchor $\mathbf{p}_2$, the center cell represents the matched tuple $(\mathbf{t}_2,\mathbf{v}_2,\mathbf{p}_2)$. The highlighted row and column contain partial-match negatives that preserve one matched pair involving the anchor and are removed by CTP. The remaining candidates are retained for contrastive normalization.}
    \label{fig:plane_loss}
\end{figure}

As shown in Eq.~\ref{eq:tensor_t2ip}, fixing one modality as the anchor leaves two sample indices free, forming a two-dimensional candidate plane in the similarity tensor. The cross-entropy loss compares the matched tuple against all $b^2$ tuples in this plane through the softmax denominator.
Among these candidates are \emph{partial-match negatives}: tuples that are negative as a whole but still contain a matched modality pair involving the anchor.

Consider a candidate plane with a fixed point-cloud anchor $\mathbf{p}_k$, where each entry $S_{ijk}$ corresponds to the tuple $(\mathbf{t}_i,\mathbf{v}_j,\mathbf{p}_k)$. The matched tuple is given by $i=j=k$. When $i=k\neq j$, the candidate is a tuple-level negative but still contains the matched text--point pair $(\mathbf{t}_k,\mathbf{p}_k)$. Likewise, when $j=k\neq i$, the candidate contains the matched image--point pair $(\mathbf{v}_k,\mathbf{p}_k)$. These candidates form the anchor-associated row and column shown in Fig.~\ref{fig:plane_loss}, and we refer to them as \emph{partial-match negatives}.

To analyze how partial-match negatives contribute to the tensor loss, consider the matched text--point similarity $a_k=\mathbf{t}_k^\top\mathbf{p}_k$ for a fixed point-cloud anchor $\mathbf{p}_k$. We define the softmax probability of candidate $(\mathbf{t}_i,\mathbf{v}_j,\mathbf{p}_k)$ within the corresponding anchor plane as
\begin{equation}
\pi_{ij}^{P}
=
\frac{
\exp\left(S_{ijk}^{\mathrm{CTP}}/\tau\right)
}{
\sum_{m=1}^{b}\sum_{n=1}^{b}
\exp\left(S_{mnk}^{\mathrm{CTP}}/\tau\right)
}.
\label{eq:point_plane_prob}
\end{equation}
Viewing the pairwise similarities in the CTP scorer as intermediate scalar variables, let $\ell_k^{P}$ denote the loss for the fixed point-cloud anchor $\mathbf{p}_k$. The matched text--point similarity $a_k$ appears in both the matched score $S_{kkk}^{\mathrm{CTP}}$ and every row-wise partial-match score $S_{kjk}^{\mathrm{CTP}}$ with $j\neq k$. Holding the remaining similarity terms fixed, the partial derivative of the loss with respect to $a_k$ is
\begin{equation}
\frac{\partial \ell_k^{P}}{\partial a_k}
=
\frac{1}{\tau}
\left(
\pi_{kk}^{P}-1
+
\sum_{j\neq k}\pi_{kj}^{P}
\right).
\label{eq:partial_match_gradient}
\end{equation}
Here, $\pi_{kk}^{P}-1$ is the contribution associated with the matched tuple, whereas $\sum_{j\neq k}\pi_{kj}^{P}$ arises from the row-wise partial-match negatives that share the same matched text--point similarity. Since the former is non-positive and the latter is non-negative, the partial-match terms partially offset the alignment signal associated with the matched tuple. The same structure arises for the matched image--point similarity and the corresponding column. A detailed derivation is provided in Appendix~\ref{app:partial_match_gradient}.

Motivated by this analysis, CTP removes the anchor-sharing partial-match negatives while retaining the matched tuple and all remaining candidates. {
We retain off-anchor diagonal candidates as negatives relative to the anchor, as additionally masking them reduces joint retrieval performance on both datasets (Appendix~\ref{sec:mask_strategy_ablation}, Table~\ref{tab:mask_strategy_comparison}).
}
\section{Experiments}
\label{sec:experiments}

\subsection{Setup}
\subsubsection{Datasets}
\label{sec:datasets}
We conduct our main experiments on object-level text--image--point-cloud triplets from nuScenes~\citep{nuscenes2019} and ULIP-ShapeNet~\citep{xue2024ulip2scalablemultimodalpretraining}, representing real-world automotive LiDAR and dense synthetic 3D shapes, respectively.

For nuScenes, we construct triplets from the official training and validation splits.
For each reference object, LiDAR observations from up to 10 sweeps are aggregated into an object-level point cloud and paired with the corresponding image crop. We use Qwen3-VL-8B-Instruct~\citep{bai2025qwen3} to generate pseudo-captions from the cropped images as textual inputs for alignment training.
The aggregated point-cloud cohort requires at least 64 unique points per object, resulting in 29,917 training and 10,711 evaluation triplets.
To assess cross-dataset generalization in real-world LiDAR, we additionally evaluate nuScenes-trained models on KITTI~\citep{Geiger2012CVPR} and the Waymo Open Dataset~\citep{Sun_2020_CVPR}, without target-domain training.
Additional details are provided in Appendix~\ref{app:dataset_construction}.

ULIP-ShapeNet contains 52,460 prepared objects across 55 categories, each associated with an 8,192-point cloud, rendered images, and captions. We construct a custom category-stratified 80/20 split, yielding 41,968 training and 10,492 evaluation objects. Both training and evaluation use the first stored image view, its first associated caption, and all 8,192 points.

\subsubsection{Implementation Details}

We use pretrained CLIP ViT-B/32 text and image encoders, where the text encoder is a Transformer-based language model~\citep{vaswani2017attention} and the image encoder is a ViT~\citep{dosovitskiy2020image}. 
For point clouds, we use a randomly initialized LitePT-S~\citep{yue2026liteptlighterstrongerpoint} encoder.
A linear projection maps the point-cloud representation into the shared 512-dimensional embedding space, and all modality embeddings are $\ell_2$-normalized before similarity computation. All three encoders are jointly optimized during training.
We train all methods for 50 epochs with a batch size of 128 using AdamW~\citep{loshchilov2017decoupled} with a weight decay
of 0.05. The initial learning rates are $10^{-5}$ for the pretrained text and image encoders and $5\times10^{-4}$ for the point encoder and projection layer, followed by a 10\% linear warm-up and cosine decay. 
Further implementation details are provided in Appendix~\ref{app:implementation}.

\subsubsection{Zero-shot Classification and Retrieval}
As shown in Fig.~\ref{pipeline}, zero-shot classification uses class prompts as text candidates and jointly compares them with the image--point-cloud representation.
For each dataset, we predefine the class vocabulary and encode one text prompt per class using the template \texttt{"This is a \{CLASS\}"}.
Given an evaluation image--point-cloud pair $(\mathbf{v}_q,\mathbf{p}_q)$, all class-text representations $\mathbf{t}_c$ are scored and the class with the highest score is selected.
For CTP and the pairwise baseline, joint classification uses
\begin{equation}
s_c
=
\frac{1}{2}\mathbf{v}_q^\top\mathbf{t}_c
+
\frac{1}{2}\mathbf{p}_q^\top\mathbf{t}_c .
\label{eq:joint_cls}
\end{equation}
For CTP, this produces the same class ranking as the full cosine-sum tuple scorer because $\mathbf{v}_q^\top\mathbf{p}_q$ is constant across candidate classes.
Symile, GRAM, and TRIANGLE are evaluated using their corresponding
native scorers.
We report overall accuracy and mean per-class accuracy (mAcc).

For joint retrieval, each evaluation sample provides a query consisting of its caption and image, $(\mathbf{t}_q,\mathbf{v}_q)$, and all point-cloud embeddings $\{\mathbf{p}_k\}$ in the fixed evaluation pool are ranked.
For CTP and the pairwise baseline, the retrieval score is
\begin{equation}
r_{qk}
=
(\mathbf{t}_q+\mathbf{v}_q)^\top\mathbf{p}_k,
\label{eq:joint_retrieval}
\end{equation}
which is equivalent in ranking to the full cosine-sum tuple scorer because $\mathbf{t}_q^\top\mathbf{v}_q$ is constant with respect to candidate $k$.
The other methods use their corresponding native scorers.
A retrieval is considered correct if at least one point cloud with the same object identity as the query appears within the top $K$ ranked candidates. We report Recall@$K$ for $K\in\{1,5,10\}$.

\subsection{Results and Analysis}
\begin{table}[!htbp]
\centering
\caption{
Zero-shot joint classification results (\%) on the nuScenes evaluation set.
Car, Truck, and Bus are merged into Vehicle.
Ped., C.V., Motor., and T.C. denote Pedestrian, Construction Vehicle,
Motorcycle, and Traffic Cone, respectively.
}
\label{tab:nuscenes10_joint_cls}
\vspace{3pt}
\resizebox{\textwidth}{!}{
\begin{tabular}{l|c|cccccccc}
\toprule
Method & mAcc & Vehicle & Ped. & Bicycle
& Trailer & C.V. & Motor. & Barrier & T.C. \\
\midrule

ULIP-style~\citep{xue2023ulip}
& 71.07 & 15.55 & 57.92 & 95.69
& 84.51 & 76.34 & 90.60 & 69.18 & 78.79 \\

Symile~\citep{saporta2024contrastingsymilesimplemodelagnostic}
& 43.11 & 21.31 & 30.07 & 1.72
& 77.10 & 55.91 & 97.44 & 50.75 & 10.61 \\

TRIANGLE~\citep{cicchetti2025triangleenablesmultimodalalignment}
& 65.75 & 17.29 & 49.27 & 93.10
& 86.20 & 78.49 & 94.87 & 48.40 & 58.33 \\

GRAM~\citep{cicchetti2025gramianmultimodalrepresentationlearning}
& 67.19 & 42.52 & 29.07 & 95.69
& 88.55 & 72.04 & 81.20 & 55.49 & 72.98 \\
\midrule

CTP (Ours)
& \textbf{74.26} & 34.19 & 61.36 & 97.41
& 85.19 & 78.49 & 97.44 & 71.53 & 68.43 \\

\bottomrule
\end{tabular}
}

\vspace{-4pt}
\end{table}
Table~\ref{tab:nuscenes10_joint_cls} reports zero-shot joint classification results on the nuScenes triplet dataset. We use a ULIP-style pairwise baseline under the same training setting as CTP: all three encoders are trainable, and the loss is defined in Eq.~\ref{eq:pair_loss}. Real-world LiDAR point clouds are sparse and often incomplete, making their alignment with image and text representations particularly challenging. The pairwise baseline achieves an mAcc of 71.07\%, outperforming Symile, GRAM, and TRIANGLE, while CTP obtains the highest mAcc of 74.26\%.

Among the joint methods, Symile achieves the lowest classification mAcc on nuScenes. Its MIP scorer is designed to capture joint statistical dependence rather than geometric co-location of modality representations, leading to a different inductive bias from CTP in the text--image--point-cloud alignment setting. GRAM and TRIANGLE achieve higher mAcc than Symile but remain below CTP in this setting. As discussed in Sec.~\ref{s_scorer}, their area- and volume-based geometric measures admit non-coincident optimum configurations, whereas the cosine-sum scorer used by CTP has a unique optimal angle configuration at multimodal co-location. However, the existence of these additional optima does not imply that training necessarily converges to such degenerate configurations. Indeed, the competitive performance of GRAM and TRIANGLE indicates that both methods can still learn meaningful multimodal alignment in the high-dimensional embedding space. We therefore interpret these additional optima as providing a weaker geometric bias toward co-location rather than constituting a failure mode. This interpretation is consistent with the moderate performance differences observed in our experiments.

\begin{table}[!htbp]
\centering
\caption{
Text--image-to-point-cloud instance retrieval (\%) on nuScenes and ULIP-ShapeNet.
}
\label{tab:joint_retrieval}

\vspace{3pt}

\small
\setlength{\tabcolsep}{7pt}

\begin{tabular}{lccc|ccc}
\toprule
\multirow{2}{*}{Method}
& \multicolumn{3}{c|}{nuScenes}
& \multicolumn{3}{c}{ULIP-ShapeNet} \\
\cmidrule(lr){2-4}\cmidrule(lr){5-7}
& R@1 & R@5 & R@10 & R@1 & R@5 & R@10 \\
\midrule

ULIP-style~\citep{xue2023ulip}
& 3.82 & 13.52 & 20.72
& 15.65 & 40.83 & 55.06 \\

Symile~\citep{saporta2024contrastingsymilesimplemodelagnostic}
& 0.49 & 2.66 & 4.92
& 3.55 & 14.74 & 25.30 \\

TRIANGLE~\citep{cicchetti2025triangleenablesmultimodalalignment}
& 6.57 & 20.04 & 30.24
& 37.70 & 68.72 & 79.61 \\

GRAM~\citep{cicchetti2025gramianmultimodalrepresentationlearning}
& 4.82 & 15.90 & 24.60
& 28.85 & 60.04 & 73.09 \\
\midrule

CTP (Ours)
& \textbf{7.44} & \textbf{22.42} & \textbf{32.18}
& \textbf{38.44} & \textbf{69.06} & \textbf{80.13} \\

\bottomrule
\end{tabular}
\end{table}
Table~\ref{tab:joint_retrieval} further evaluates instance-level multimodal alignment through text--image-to-point-cloud retrieval. On nuScenes, CTP achieves the highest retrieval performance among the compared methods, reaching 7.44\% R@1, 22.42\% R@5, and 32.18\% R@10.

A larger separation between pairwise and joint geometric methods is observed on ULIP-ShapeNet. The pairwise baseline obtains 15.65\% R@1, 40.83\% R@5, and 55.06\% R@10, whereas GRAM, TRIANGLE, and CTP all substantially improve instance retrieval. CTP achieves the best performance, reaching 38.44\% R@1, 69.06\% R@5, and 80.13\% R@10. In contrast, Symile remains below the pairwise baseline on both datasets, consistent with its different statistical rather than geometric notion of multimodal similarity.

Among the geometric joint methods, the retrieval performance follows the same ordering on both datasets: CTP, TRIANGLE, and GRAM. This trend is consistent with the geometric analysis. The cosine-sum scorer used by CTP uniquely favors co-location of the three modality embeddings, while the triangle-area scorer admits additional degenerate configurations and the Gram-volume scorer has an even broader set of zero-volume configurations corresponding to linearly dependent embeddings.

\begin{table}[!htbp]
\centering
\small
\setlength{\tabcolsep}{5pt}

\caption{
Cross-dataset joint classification mAcc (\%) on KITTI and Waymo
after training on nuScenes.}
\label{tab:joint_unseen}

\vspace{3pt}

\begin{tabular}{lcc}
\toprule
Method & KITTI & Waymo \\
\midrule
ULIP-style~\citep{xue2023ulip}
    & 50.59 & 57.76 \\
Symile~\citep{saporta2024contrastingsymilesimplemodelagnostic}
    & 34.87 & 42.67 \\
TRIANGLE~\citep{cicchetti2025triangleenablesmultimodalalignment}
    & 60.47 & 46.71 \\
GRAM~\citep{cicchetti2025gramianmultimodalrepresentationlearning}
    & 52.47 & 49.26 \\
\midrule
CTP (Ours)
    & \textbf{64.25} & \textbf{65.07} \\
\bottomrule
\end{tabular}

\end{table}

The relative gains of the joint geometric methods are also larger on ULIP-ShapeNet than on nuScenes. This observation is consistent with the hypothesis that richer 3D point clouds facilitate joint multimodal alignment, whereas sparse and incomplete LiDAR point clouds make the alignment problem more challenging and reduce the performance gap between pairwise and joint methods.

To evaluate cross-dataset generalization, we directly apply the nuScenes-trained models to KITTI and Waymo without target-domain training or adaptation. Both evaluation sets are constructed following the same object-level text--image--point-cloud triplet pipeline as nuScenes.

As shown in Table~\ref{tab:joint_unseen}, all compared methods retain zero-shot classification capability under this domain shift. Among them, CTP achieves the highest mAcc on both datasets, reaching 64.25\% on KITTI and 65.07\% on Waymo. These results show that the multimodal representations learned on nuScenes remain effective across different real-world LiDAR domains. The consistent performance of CTP on both the denser KITTI point clouds and the sparser Waymo point clouds further suggests robustness to variations in 3D point-cloud density and dataset distribution. Overall, the observed cross-dataset transferability supports the potential of the learned point-cloud encoder as a general 3D representation backbone for downstream multimodal applications.

\begin{figure}[!htbp]
    \centering
    \includegraphics[width=0.65\linewidth]{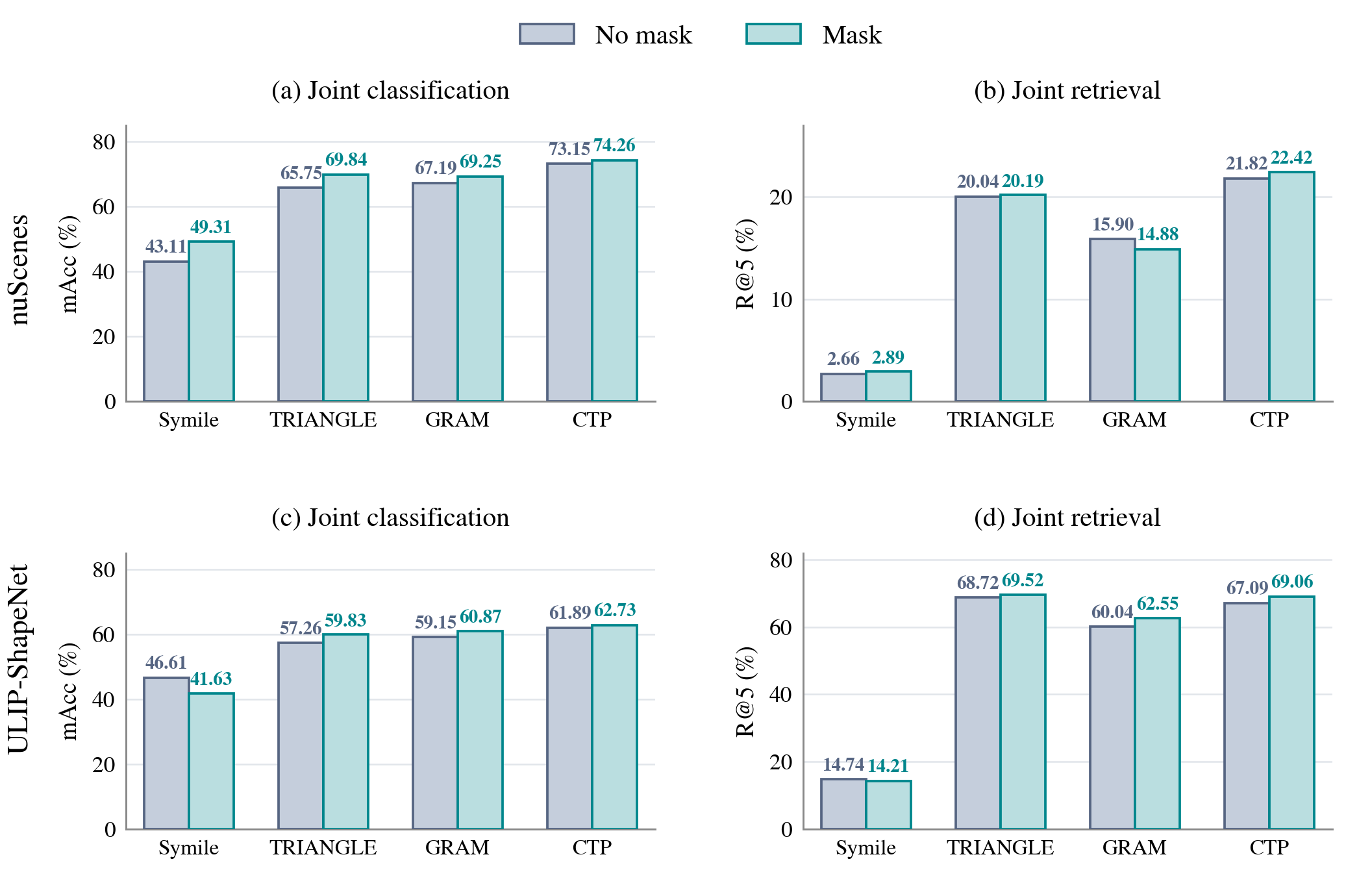}
    \caption{
    Effect of masking partial-match negatives on zero-shot joint classification mAcc (\%) {and joint retrieval R@5 (\%)} on nuScenes and ULIP-ShapeNet.
    No mask uses all tuple candidates, whereas Mask removes anchor-sharing partial-match negatives.
    }
    \label{fig:mask_effect}
\end{figure}

Figure~\ref{fig:mask_effect} compares joint classification {and retrieval} performance with and without masking partial-match negatives.
For the geometric joint scorers, TRIANGLE, GRAM, and CTP, masking consistently improves mAcc on both nuScenes and ULIP-ShapeNet.
This observation supports the motivation for excluding partial-match negatives, which are negative at the tuple level but still preserve a valid modality pair involving the anchor.
{
For joint retrieval, masking improves R@5 for TRIANGLE and CTP on both datasets, while GRAM improves on ULIP-ShapeNet but exhibits a decrease on nuScenes.
}

In contrast, the effect of masking on Symile is less consistent
across datasets,
{for both classification and retrieval,}
suggesting that the influence of partial-match negatives depends on the underlying tuple similarity function. Overall, these results indicate that masking is particularly beneficial for the geometric joint alignment objectives considered in this work.
{
The retrieval results further highlight that these benefits depend on the scorer and dataset, with CTP consistently improving both mAcc and R@5 across the two datasets.
}

\section{Conclusion}
\label{sec:conclusion}

We presented CTP for joint contrastive alignment of text, image, and point-cloud representations. Our experiments on nuScenes and ULIP-ShapeNet show that CTP achieves strong category-level alignment and instance-level cross-modal retrieval compared with a matched pairwise baseline. We further show that the cosine-sum scorer provides a direct geometric bias toward multimodal co-location, while alternative area- and volume-based scorers admit additional non-coincident optima. Finally, we identify partial-match negatives in joint contrastive learning and find that masking them consistently improves the performance of the geometric joint objectives considered in this work. These results highlight the importance of scorer geometry and candidate construction for joint language--image--3D representation learning.

\bibliography{references}
\bibliographystyle{references}

\appendix
\section{Triplet Dataset}
\label{app:dataset_construction}

\subsection{Triplet Construction}
\label{app:triplet_construction}
\begin{figure}[!htbp]
    \centering
    \includegraphics[width=1\linewidth]{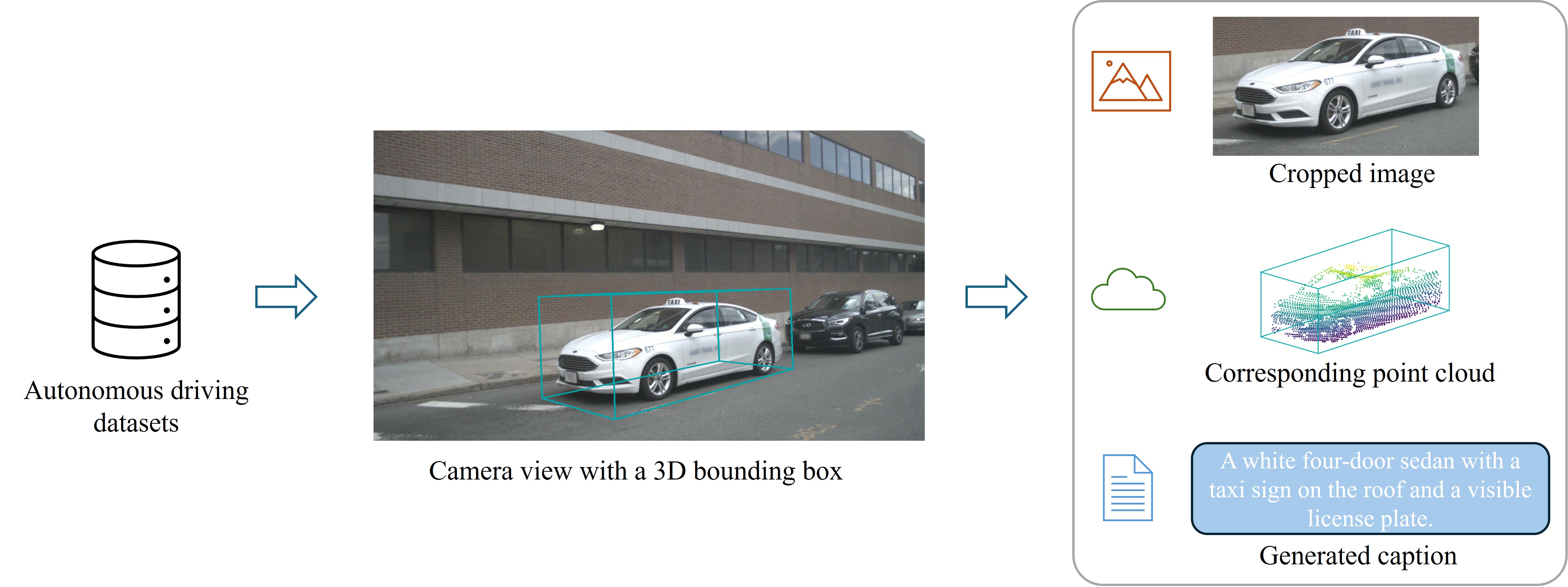}
    \caption{Overview of object-level triplet construction using a nuScenes example. For each annotated object, we extract the corresponding image crop and LiDAR point cloud and generate a pseudo-caption from the cropped image and object category. The point cloud shown here aggregates up to 10 LiDAR sweeps.}
    \label{data_pipeline}
\end{figure}
We construct object-level text--image--point-cloud triplets from nuScenes~\citep{nuscenes2019}, KITTI~\citep{Geiger2012CVPR}, and the Waymo Open Dataset~\citep{Sun_2020_CVPR}. As shown in Fig.~\ref{data_pipeline}, each annotated object is associated with a point-cloud segment, a corresponding image crop, and a textual description. The 3D annotations and camera--LiDAR correspondences establish the association between the image and point-cloud observations, while image-conditioned pseudo-captioning provides the textual modality.

For nuScenes, we construct the training and evaluation sets from the official training and validation scenes, respectively. Reference frames are selected at 2-second intervals. For each reference frame, we aggregate the current LiDAR scan and nine preceding scans. Using sensor calibration and ego poses, all scans are transformed into the reference ego coordinate system. We then extract the points inside each annotated 3D bounding box at the reference frame.

To reduce points near the bottom of an object box, we transform the extracted points into the box-local coordinate system using the full box rotation and retain points satisfying
\begin{equation}
z_{\mathrm{local}} \geq -\frac{h}{2} + \delta,
\qquad \delta = 0.1\,\mathrm{m},
\end{equation}
where $h$ denotes the box height. This operation removes a thin region at the box bottom. The final nuScenes cohort retains objects for which all ten scans are available and that contain at least 64 unique XYZ points after filtering.

For the image modality, we retain nuScenes annotations with visibility tokens of at least 2 and project each object's 3D bounding box into the available camera views. A view is considered valid when at least 80\% of its projected 2D bounding rectangle lies inside the image. Among valid views, we select the one with the largest in-image projected area, using proximity to the image center to break ties. We extract the corresponding image crop with a 5-pixel margin, clipped to the image boundaries. The annotation visibility criterion and the image-boundary criterion are applied separately.

\begin{figure}[!htbp]
    \centering
    \includegraphics[width=0.85\linewidth]{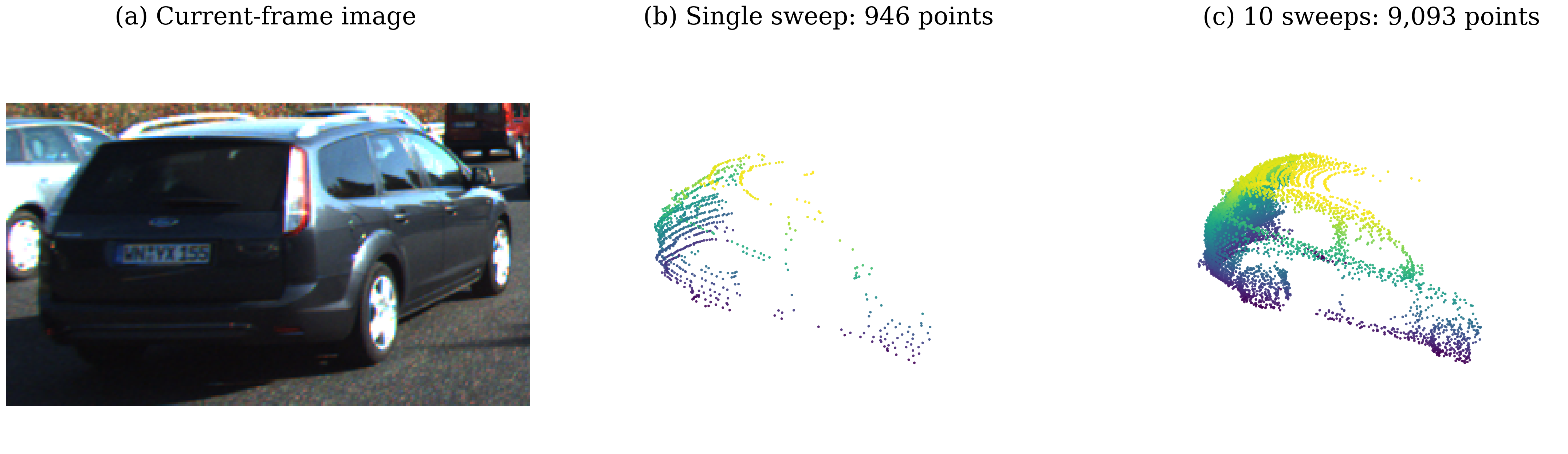}
    \caption{Temporal aggregation in KITTI. The single-frame object point cloud is substantially densified by aggregating up to 10 frames of the same tracked object.}
    \label{fig:kitti_sweeps_example}
\end{figure}

KITTI follows the same general triplet-construction procedure as nuScenes, but uses annotated track identities and frame-specific 3D boxes to align the same object across frames before aggregation. Figure~\ref{fig:kitti_sweeps_example} shows the effect of this temporal aggregation. We apply the same box-bottom exclusion and image-cropping settings as for nuScenes, and retain aggregated point clouds containing at least 50 points.

Waymo differs in that we use single-frame object point clouds rather than temporal aggregation. LiDAR boxes are associated with projected camera boxes using the segment identifier, frame timestamp, and object identifier. We retain object point clouds containing at least 15 points. If an object is visible in multiple cameras within the same frame, each valid camera crop forms a separate triplet.

We use Qwen3-VL-8B-Instruct~\citep{bai2025qwen3} to generate descriptive pseudo-captions for the nuScenes, KITTI, and Waymo triplet datasets. For each object, the captioning model is provided with the cropped RGB image and its object category. The same caption-generation procedure is applied to all three datasets.

For reproducibility, we provide the system and user prompts used for caption generation below. In the user prompt, \texttt{\{category\}} is replaced by the corresponding object label.

\begin{quote}
\small
\textbf{System prompt:}

You generate one concise factual sentence about the specified road object. Follow every constraint exactly and do not add scene context.

\medskip
\textbf{User prompt:}

Target object category: \texttt{\{category\}}.

Describe only the target object in one factual sentence.
Describe its visible shape, approximate scale, orientation, and distinctive structure.
Mention color only when it is clearly visible.
Do not mention the image, photo, scene, background, camera, or crop.
Do not infer motion, speed, intent, or hidden attributes.
If an attribute is uncertain, omit it.
\end{quote}

\subsection{Dataset Statistics}
\label{app:dataset_statistics}

\begin{figure}[!htbp]
    \centering
    \includegraphics[width=1\linewidth]{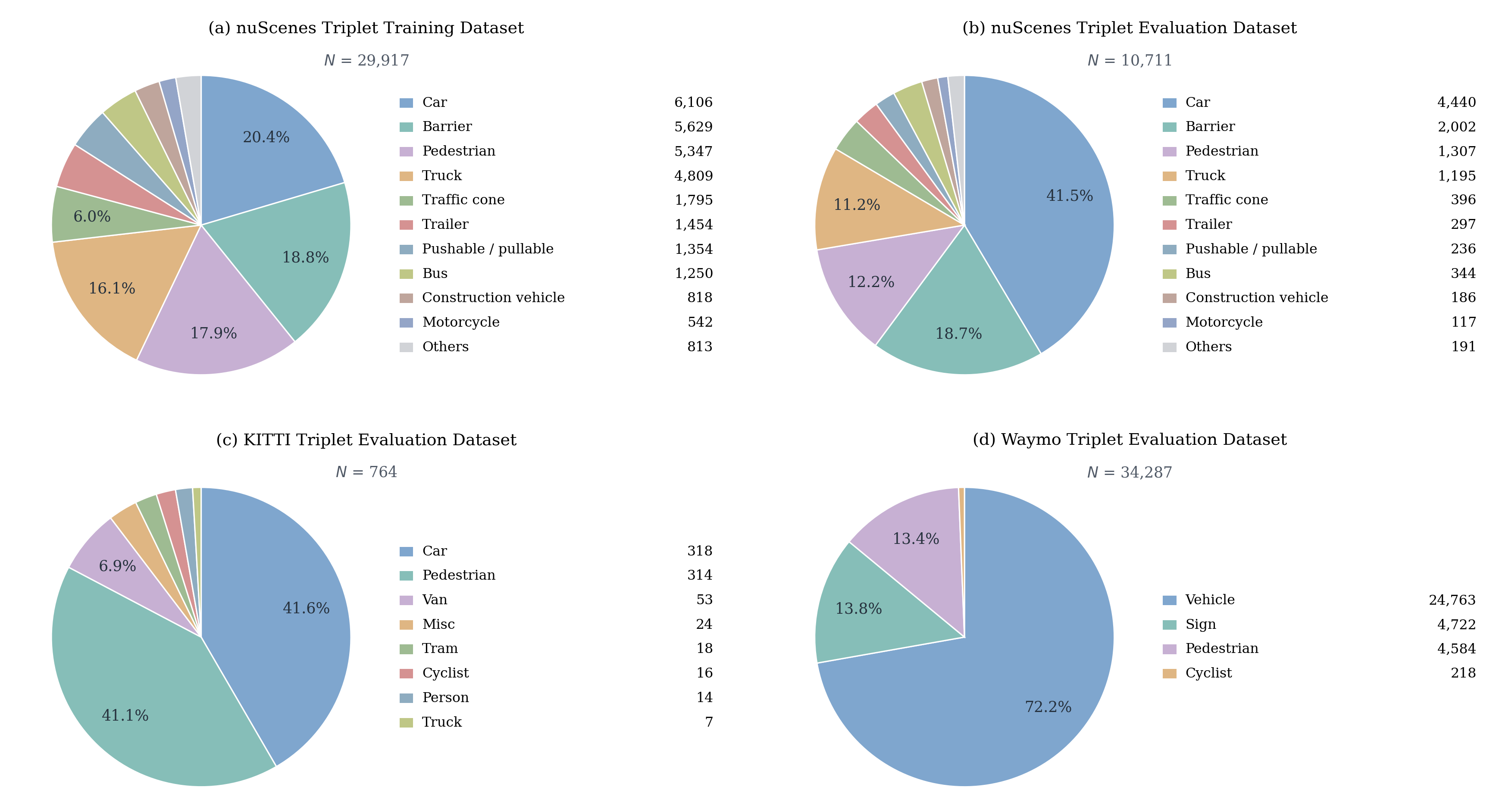}
    \caption{Category distributions of the constructed triplet datasets: 
    (a) nuScenes training, (b) nuScenes evaluation, (c) KITTI evaluation, 
    and (d) Waymo evaluation. $N$ denotes the number of triplets, and 
    legend values indicate category counts. For nuScenes, ``Others'' groups 
    the Bicycle, Bicycle Rack, Debris, Emergency, and Animal categories for 
    visualization.}
    \label{fig:dataset_distribution}
\end{figure}

Figure~\ref{fig:dataset_distribution} summarizes the category distributions of the autonomous-driving triplet datasets. The nuScenes dataset contains 29,917 training triplets and 10,711 evaluation triplets. For cross-dataset evaluation, the KITTI subset contains 764 triplets across eight categories, and the Waymo subset contains 34,287 triplets across four categories.

\subsection{Representative Triplets}
\label{app:triplet_examples}

\begin{figure}[!htbp]
    \centering
    \includegraphics[width=1\linewidth]{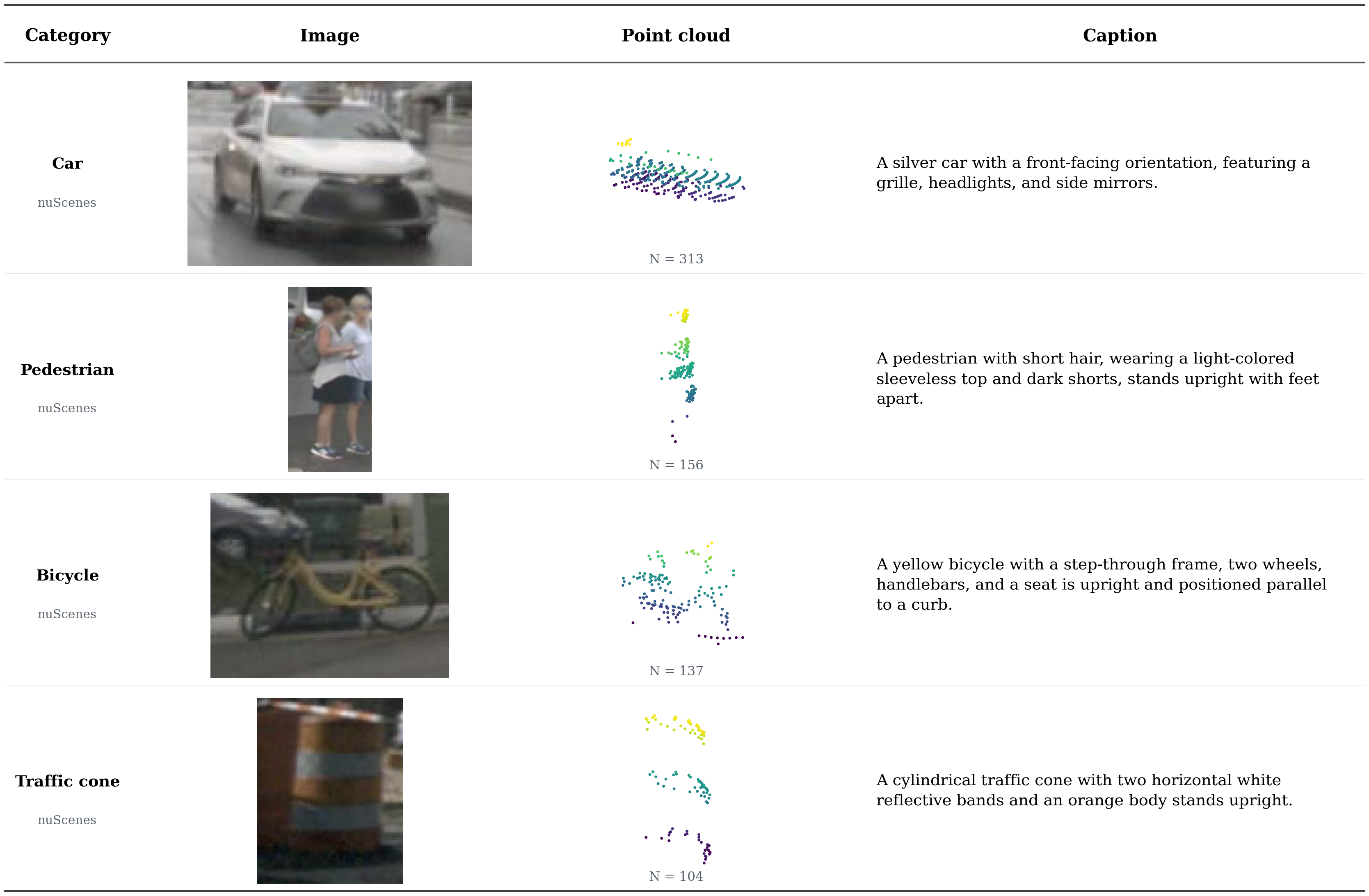}
    \caption{Selected triplets from the nuScenes evaluation set. Each row shows the object category, corresponding image crop, point cloud, and pseudo-caption. $N$ denotes the number of saved points.}
    \label{fig:nuscenes_examples}
\end{figure}

\begin{figure}[!htbp]
    \centering
    \includegraphics[width=1\linewidth]{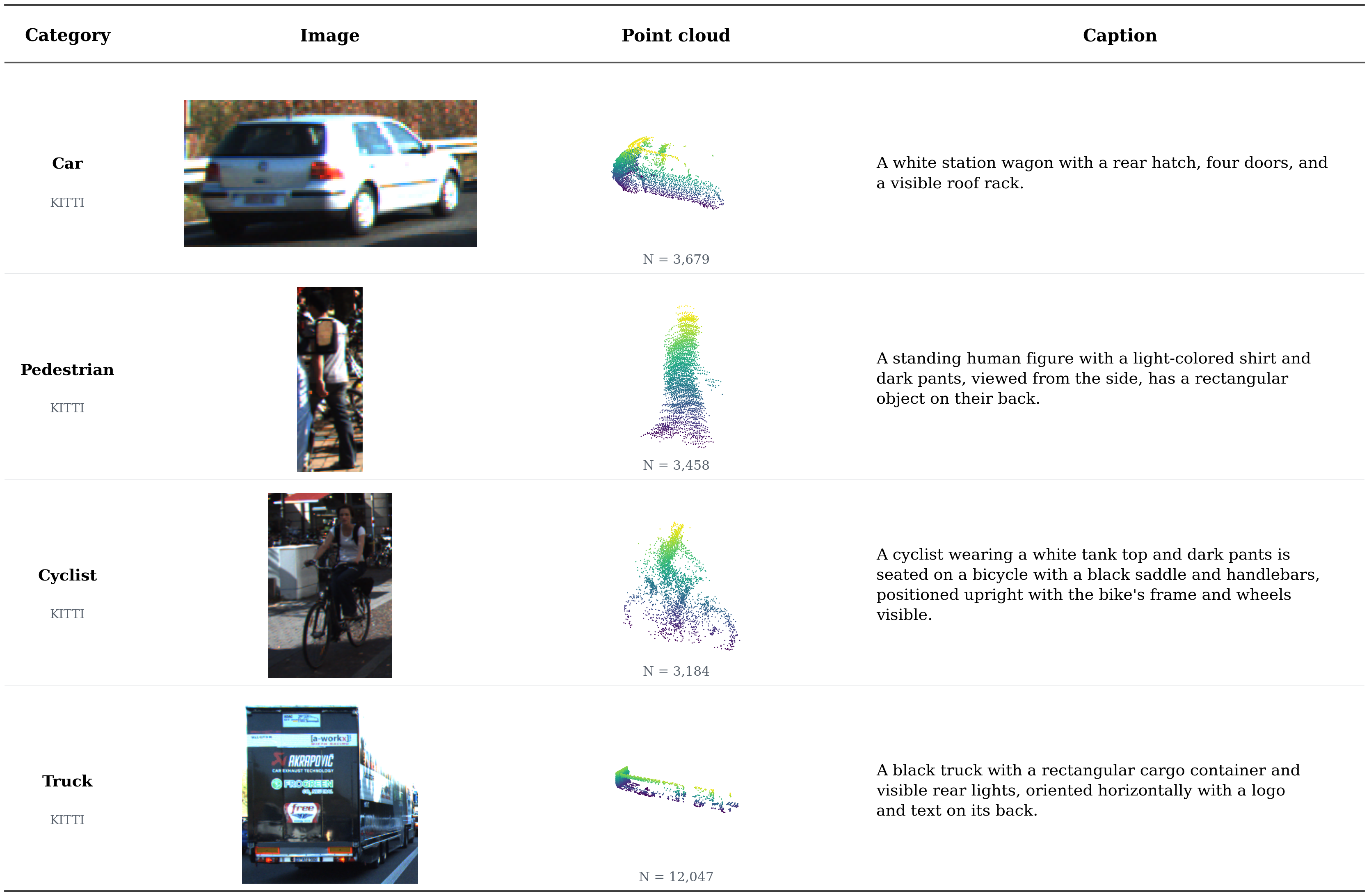}
    \caption{Selected triplets from the KITTI evaluation set. Each row shows the object category, corresponding image crop, point cloud, and pseudo-caption. $N$ denotes the number of saved points.}
    \label{fig:kitti_examples}
\end{figure}

\begin{figure}[!htbp]
    \centering
    \includegraphics[width=1\linewidth]{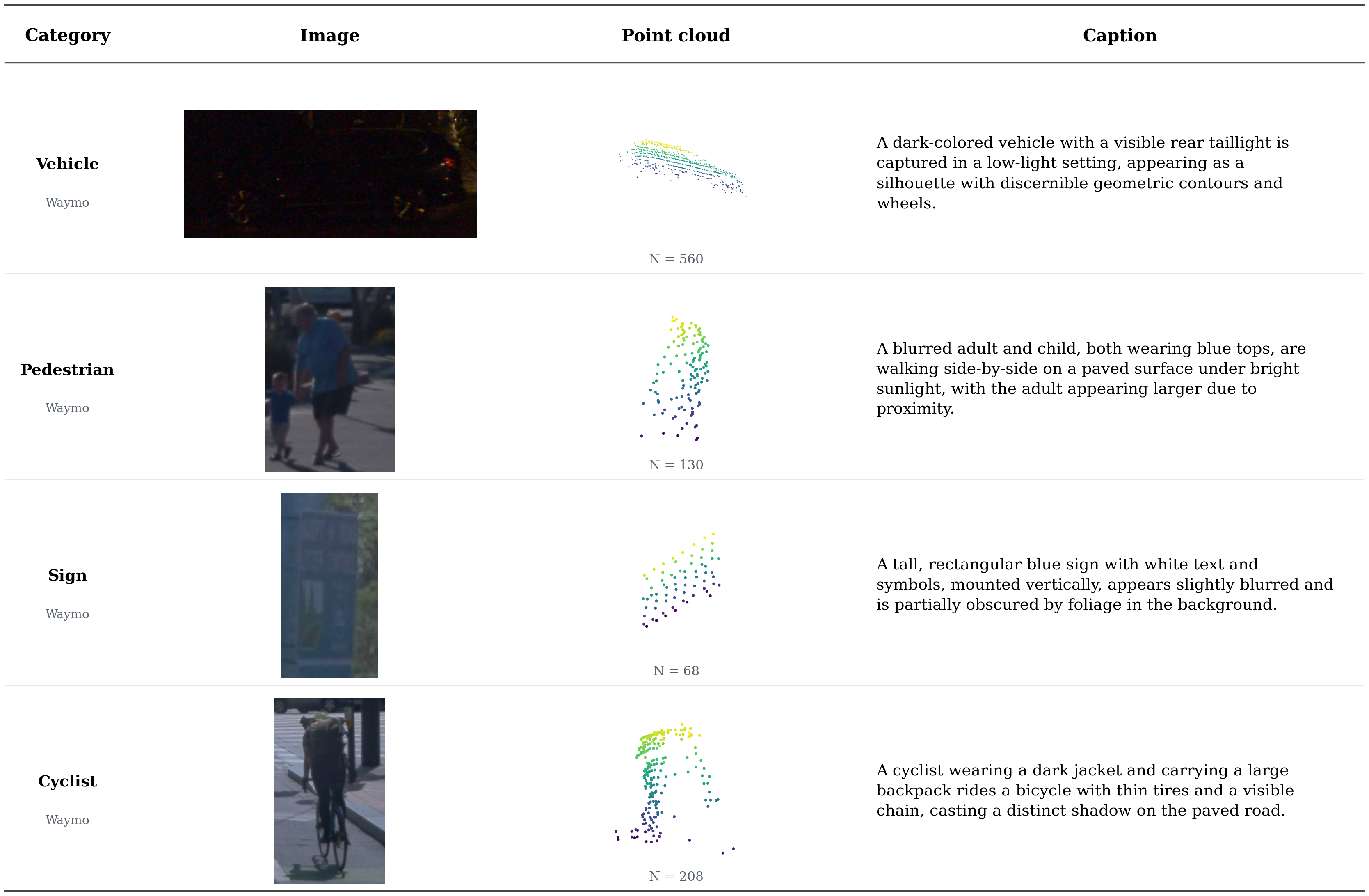}
    \caption{Selected triplets from the Waymo evaluation set. Each row shows the object category, corresponding image crop, point cloud, and pseudo-caption. $N$ denotes the number of saved points.}
    \label{fig:waymo_examples}
\end{figure}

Figures~\ref{fig:nuscenes_examples}, \ref{fig:kitti_examples}, and \ref{fig:waymo_examples} present selected triplets from the nuScenes, KITTI, and Waymo evaluation sets, respectively. These examples qualitatively show differences in point-cloud density and geometric coverage across the datasets. The selected nuScenes point clouds remain relatively sparse and incomplete despite temporal aggregation, whereas the KITTI examples exhibit denser and more complete geometric coverage. Waymo uses single-frame point clouds and, in the selected examples, shows sparser coverage than KITTI.

\section{Implementation Details}
\label{app:implementation}

Each cropped RGB image is resized with bilinear interpolation so that its longer side is 224 pixels while preserving the aspect ratio, and is then centered on a black $224\times224$ canvas. Images are normalized using the CLIP channel statistics. The same deterministic preprocessing is used for training and evaluation. Text is tokenized using the CLIP tokenizer with a context length of 77 tokens, with longer descriptions truncated.

The point-cloud encoder uses only XYZ coordinates. For the driving datasets, each object point cloud is centered by subtracting its centroid while retaining its metric scale. ULIP-ShapeNet uses the normalized coordinates provided by the prepared assets. All stored points are retained without fixed-size sampling. Before entering LitePT-S, points within each voxel are averaged using a grid size of 0.05 in metric coordinates for the driving datasets and normalized coordinates for ULIP-ShapeNet. The encoded point features are mean-pooled before the linear projection into the shared embedding space.

Table~\ref{tab:implementation_details} summarizes the main training configuration. All compared methods use the same input processing, encoder architectures, initialization scheme, and optimization settings. Unless otherwise stated, reported results use the final checkpoint after 50 epochs.

\begin{table}[!htbp]
\centering
\caption{Implementation and training configuration used in the main experiments.}
\label{tab:implementation_details}
\small
\begin{tabular}{ll}
\toprule
Setting & Configuration \\
\midrule
Text encoder & CLIP ViT-B/32, pretrained and trainable \\
Image encoder & CLIP ViT-B/32, pretrained and trainable \\
Point encoder & LitePT-S, randomly initialized and trainable \\
Point projection & Linear, $504\rightarrow512$, with bias \\
Embedding dimension & 512 \\
Embedding normalization & $\ell_2$ normalization \\
Training epochs & 50 \\
Global batch size & 128 \\
Optimizer & AdamW \\
Weight decay & 0.05 \\
CLIP learning rate & $1\times10^{-5}$ \\
Point/projector learning rate & $5\times10^{-4}$ \\
LR schedule & 10\% linear warm-up + cosine decay \\
Gradient clipping & Global norm 1.0 \\
Mixed precision & BF16 encoder forward, FP32 scoring/loss \\
Logit-scale initialization & $1/0.07$ \\
Logit-scale cap & 1000 \\
Random seed & 42 \\
\bottomrule
\end{tabular}
\end{table}
\section{Additional Theoretical Analysis}
\label{app:theory}

\subsection{Alternative Tuple Scorers}
\label{app:scorer_analysis}
Different multimodal tuple scorers impose different inductive biases. Symile~\citep{saporta2024contrastingsymilesimplemodelagnostic}, for example, uses the multilinear inner product (MIP),
\begin{equation}
S_{\mathrm{MIP}}(\mathbf{t},\mathbf{v},\mathbf{p}) = \sum_{r=1}^{d} t_r v_r p_r .
\label{eq:mip_scorer}
\end{equation}
Unlike the CTP scorer, MIP is not determined solely by the three pairwise angles $(\alpha,\beta,\gamma)$. A common rotation of the three embeddings preserves all three pairwise angles but can change the MIP value. It therefore captures a different notion of multimodal similarity from the angle-based geometric co-location considered by CTP.

We further consider the geometric properties of CTP and two alternatives inspired by TRIANGLE~\citep{cicchetti2025triangleenablesmultimodalalignment} and GRAM~\citep{cicchetti2025gramianmultimodalrepresentationlearning}. Let
\begin{equation}
a=\cos\alpha,\qquad b=\cos\beta,\qquad c=\cos\gamma.
\label{eq:abc_cos}
\end{equation}
For unit-normalized embeddings, the CTP scorer satisfies $S_{\mathrm{CTP}}=a+b+c$. The total squared pairwise distance is therefore
\begin{equation}
D^2 = \|\mathbf{t}-\mathbf{v}\|_2^2 + \|\mathbf{t}-\mathbf{p}\|_2^2 + \|\mathbf{v}-\mathbf{p}\|_2^2 = 6-2(a+b+c)=6-2S_{\mathrm{CTP}}.
\label{eq:ctp_distance}
\end{equation}
Thus, maximizing the CTP scorer is equivalent to minimizing the total squared pairwise distance.

TRIANGLE measures the area of the triangle formed by the three embedding endpoints. Its squared area can be written as
\begin{equation}
A^2 = \frac{1}{4}\left[3-2(a+b+c)+2(ab+ac+bc)-(a^2+b^2+c^2)\right].
\label{eq:triangle_area}
\end{equation}
GRAM is based on the volume spanned by the three embedding vectors, whose squared volume is given by the determinant of their Gram matrix,
\begin{equation}
V^2 = 1+2abc-a^2-b^2-c^2.
\label{eq:gram_volume}
\end{equation}
In training, TRIANGLE and GRAM use the negative area and negative volume, respectively, as tuple scores, so maximizing the contrastive score corresponds to minimizing the geometric measure. Specifically, we use $s_{\mathrm{TRIANGLE}}=-\sqrt{\max(A^2,0)+\epsilon}$ and $s_{\mathrm{GRAM}}=-\sqrt{\max(V^2,0)+\epsilon}$ with $\epsilon=10^{-8}$ for numerical stability.

\begin{figure}[!htbp]
    \centering
    \includegraphics[width=0.85\linewidth]{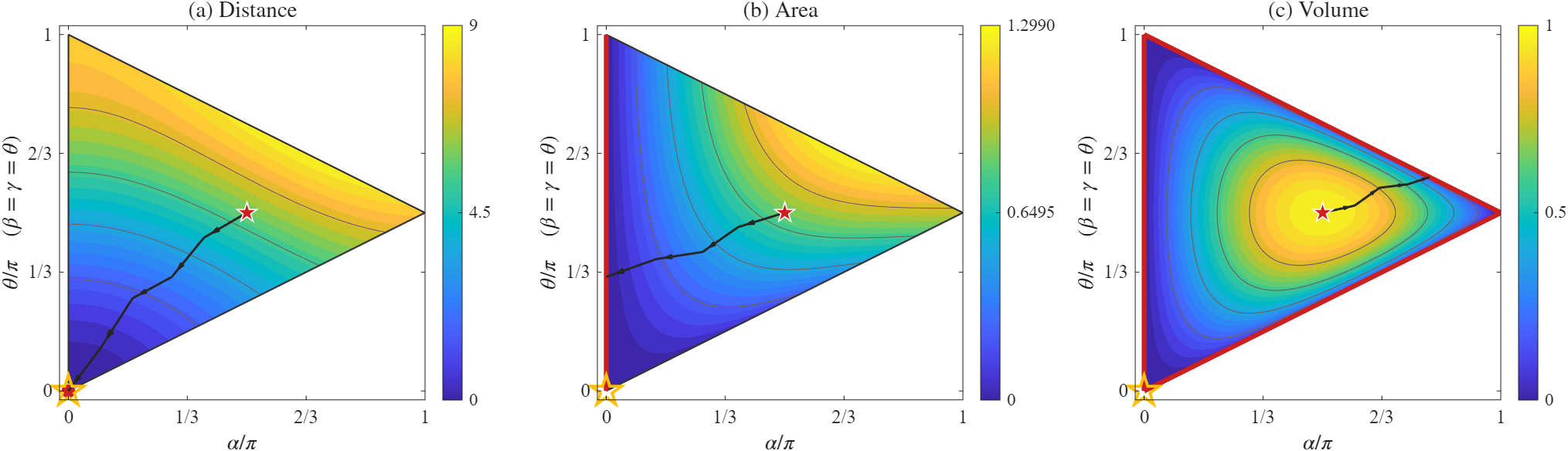}
    \caption{Geometric tuple measures on the $\beta=\gamma=\theta$ slice of the triplet angle space in Fig.~\ref{fig:angle_space}. The yellow star denotes multimodal co-location, red lines denote additional minima, and black curves illustrate possible directions of decreasing measure. The color bars indicate the value of the corresponding geometric measure at each angle configuration.}
    \label{fig:scorer}
\end{figure}

Figure~\ref{fig:scorer} provides a two-dimensional view of the triplet angle space by visualizing the light-purple plane in Fig.~\ref{fig:angle_space}, corresponding to the slice $\beta=\gamma=\theta$. Our desired alignment configuration is the co-location of all three embeddings, corresponding to $(\alpha,\theta)=(0,0)$ and marked by the yellow star. The red lines indicate additional configurations at which the corresponding geometric measure attains its minimum. Starting from an illustrative configuration marked by the red star, the black curves show possible directions of decreasing measure. For the total squared pairwise distance, the unique minimum is the co-location configuration, so decreasing the measure leads toward the yellow star. In contrast, area and volume admit additional non-coincident minima, allowing the measure to decrease toward configurations in which the three embeddings are not co-located.

These properties show the different inductive biases of the tuple scorers considered in our experiments. The cosine-sum scorer directly favors pairwise co-location, whereas the area- and volume-based scorers also admit non-coincident optima. Their empirical behavior is compared in Sec.~\ref{sec:experiments}.

\subsection{Gradient Analysis of Partial-Match Negatives}
\label{app:partial_match_gradient}

We provide the derivation of Eq.~\ref{eq:partial_match_gradient}. Consider a fixed point-cloud anchor $\mathbf{p}_k$. The corresponding per-anchor tensor loss is
\begin{equation}
\ell_k^{P}
=
-\log
\frac{
\exp\left(S_{kkk}^{\mathrm{CTP}}/\tau\right)
}{
\sum_{i=1}^{b}\sum_{j=1}^{b}
\exp\left(S_{ijk}^{\mathrm{CTP}}/\tau\right)
}
=
-\frac{S_{kkk}^{\mathrm{CTP}}}{\tau}
+\log Z_k^{P},
\label{eq:appendix_point_loss}
\end{equation}
where
\begin{equation}
Z_k^{P}
=
\sum_{i=1}^{b}\sum_{j=1}^{b}
\exp\left(S_{ijk}^{\mathrm{CTP}}/\tau\right).
\end{equation}

Let
\begin{equation}
a_k=\mathbf{t}_k^\top\mathbf{p}_k
\end{equation}
denote the matched text--point similarity. Under the additive CTP scorer,
\begin{equation}
S_{ijk}^{\mathrm{CTP}}
=
\mathbf{t}_i^\top\mathbf{v}_j
+
\mathbf{t}_i^\top\mathbf{p}_k
+
\mathbf{v}_j^\top\mathbf{p}_k,
\end{equation}
the term $a_k$ appears in every score with $i=k$, including the matched tuple $S_{kkk}^{\mathrm{CTP}}$ and the row-wise partial-match negatives $S_{kjk}^{\mathrm{CTP}}$ for $j\neq k$. Treating the pairwise similarities as intermediate scalar variables and holding the remaining terms fixed,
\begin{equation}
\frac{\partial S_{ijk}^{\mathrm{CTP}}}{\partial a_k}
=
\begin{cases}
1, & i=k,\\
0, & i\neq k.
\end{cases}
\label{eq:appendix_score_derivative}
\end{equation}

Differentiating Eq.~\ref{eq:appendix_point_loss} with respect to $a_k$ gives
\begin{align}
\frac{\partial \ell_k^{P}}{\partial a_k}
&=
-\frac{1}{\tau}
+
\frac{1}{\tau Z_k^{P}}
\sum_{j=1}^{b}
\exp\left(S_{kjk}^{\mathrm{CTP}}/\tau\right) \\
&=
\frac{1}{\tau}
\left(
-1+\sum_{j=1}^{b}\pi_{kj}^{P}
\right),
\end{align}
where
\begin{equation}
\pi_{ij}^{P}
=
\frac{
\exp\left(S_{ijk}^{\mathrm{CTP}}/\tau\right)
}{
Z_k^{P}
}
\end{equation}
is the softmax probability of candidate $(\mathbf{t}_i,\mathbf{v}_j,\mathbf{p}_k)$ within the point-cloud-anchored plane. Separating the matched tuple from the row-wise partial-match negatives yields
\begin{equation}
\frac{\partial \ell_k^{P}}{\partial a_k}
=
\frac{1}{\tau}
\left(
\pi_{kk}^{P}-1
+
\sum_{j\neq k}\pi_{kj}^{P}
\right),
\end{equation}
which gives Eq.~\ref{eq:partial_match_gradient}. The term $\pi_{kk}^{P}-1$ is associated with the matched tuple, whereas the nonnegative summation arises from partial-match negatives that contain the same matched text--point similarity $a_k$ and therefore partially offsets this contribution.

An analogous derivation applies to the matched image--point similarity $\mathbf{v}_k^\top\mathbf{p}_k$, for which the corresponding partial-match contribution arises from the column $j=k$.

\section{Additional Results}
\label{app:additional_results}
\subsection{Effect of the Number of Sampled Candidates}
\label{app:candidate_sampling}

\begin{figure}[!htbp]
    \centering
    \includegraphics[width=0.62\linewidth]{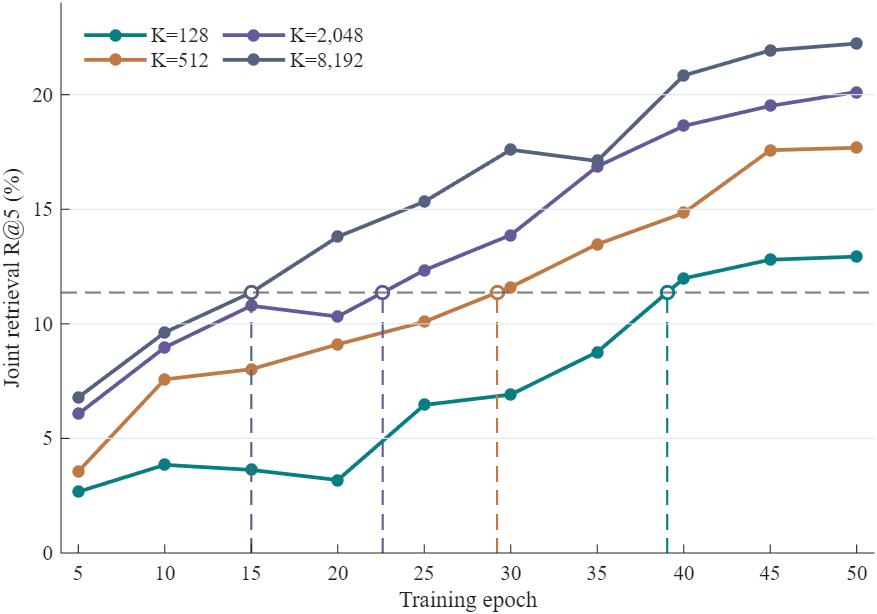}
    \caption{Effect of the training candidate count $K$ on nuScenes joint text--image-to-point-cloud retrieval. Dashed lines mark a shared R@5 level of 11.365\%.}
    \label{fig:candidate_sampling}
\end{figure}

As discussed in Fig.~\ref{fig:plane_loss}, each anchor is contrasted against $b^2$ tuple candidates. Row-and-column masking leaves $b^2-2b+2$ candidates, including the positive tuple. We retain the positive tuple and randomly sample $K-1$ negatives from the remaining set, yielding $K\in\{128,512,2048,8192\}$ candidates per anchor. All runs use a batch size of 128, with other training settings unchanged from the main experiments. Joint text--image-to-point-cloud retrieval is evaluated every five epochs over a fixed pool instances.

Figure~\ref{fig:candidate_sampling} shows that larger candidate sets generally yield higher R@5, particularly in later epochs. However, smaller candidate sets can reach the same intermediate retrieval level with additional training epochs. The dashed guides show this at R@5 = 11.365\%: $K=8192$ reaches this level at epoch 15, while $K=2048$, $512$, and $128$ reach it at approximately epochs 23, 29, and 39, respectively, using linear interpolation between evaluations. Candidate sampling therefore offers a trade-off between the number of candidates per anchor and the training epochs needed to reach this retrieval level, although smaller candidate sets still achieve lower R@5 at epoch 50.
{

\subsection{Comparison under Matched Candidate Counts}
\label{sec:matched_candidates}
Figure~\ref{fig:matched_candidates} compares ULIP-style and CTP on nuScenes over 100 training epochs.
Both methods use the same encoders, batch size of 128, and training schedule.
Each pairwise contrastive term uses 127 negatives per anchor, while CTP samples 127 negative triplets per anchor, giving 128 candidates including the positive.
ULIP-style achieves higher retrieval recall early in training, whereas CTP improves more gradually and eventually surpasses it across all three retrieval metrics.
These results suggest that sampling from CTP's larger pool of multimodal negative combinations can benefit alignment, even when the number of candidates used per contrastive term is matched.
At epoch 100, CTP achieves 86.54\% mAcc and 4.21/14.64/23.22\% R@1/R@5/R@10, compared with 71.42\% and 3.84/12.57/19.57\% for ULIP-style.

\begin{figure}[!htbp]
    \centering
    \includegraphics[width=1\linewidth]{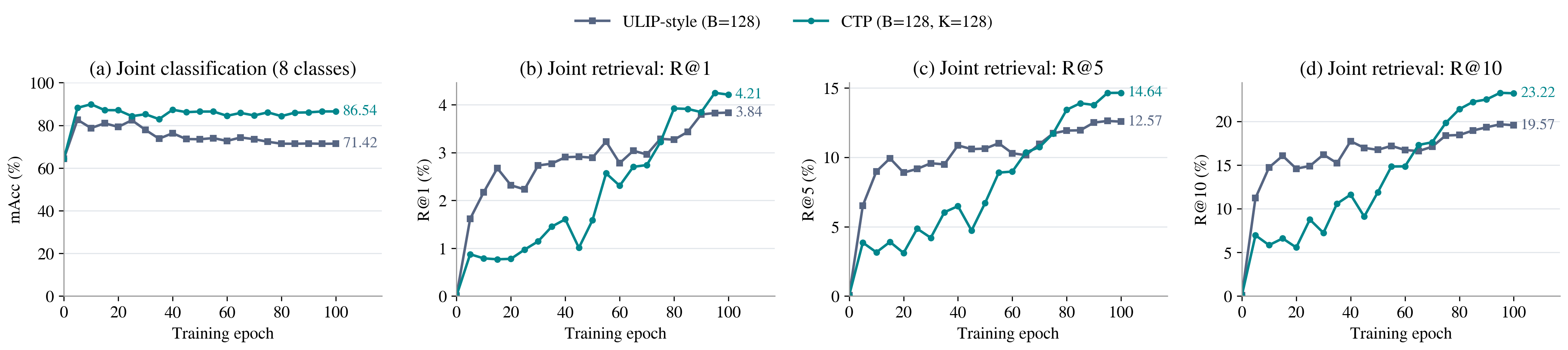}
    \caption{Training curves for ULIP-style and CTP with matched
    per-term candidate counts on nuScenes.}
    \label{fig:matched_candidates}
\end{figure}

\subsection{Comparison of Captioning Strategies}
\label{sec:caption_comparison}

Figure~\ref{fig:caption} compares category-based and category-agnostic captions on nuScenes.
Category-based captions are generated with ground-truth category guidance, whereas category-agnostic captions omit this guidance.
Each method is trained using the corresponding training captions, and joint retrieval uses the corresponding evaluation captions.
CTP achieves the best joint classification and retrieval results under both settings. 
Without category guidance, its eight-class mAcc decreases from 74.26\% to 60.63\%, while R@5 decreases from 22.42\% to 21.12\%. 
These results show that CTP's relative advantage persists without category guidance, although classification is more sensitive to the captioning strategy.

\begin{figure}[!htbp]
    \centering
    \includegraphics[width=1\linewidth]{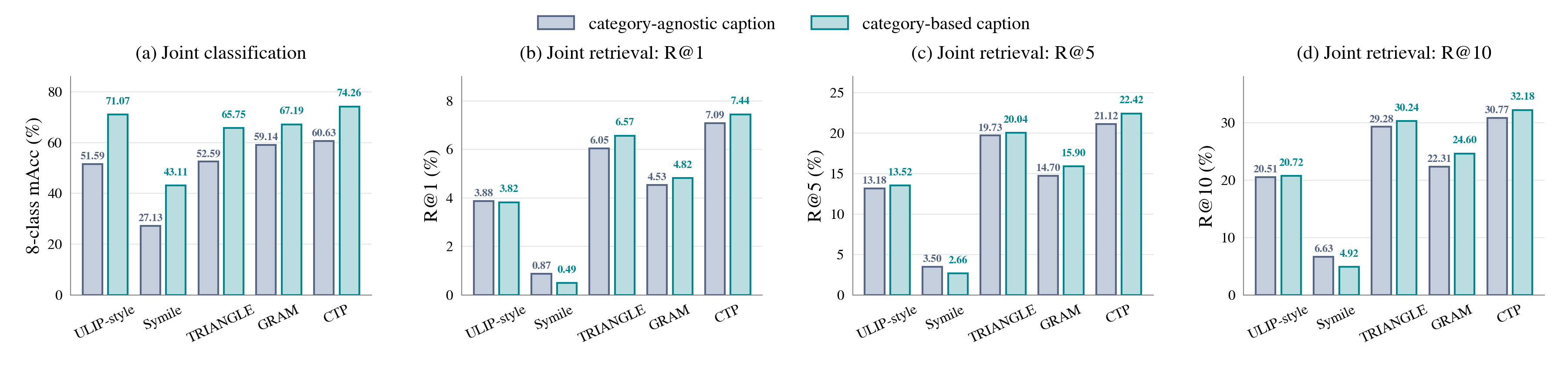}
    \caption{Comparison of category-based and category-agnostic captions
    on nuScenes for eight-class joint classification mAcc (\%)
    and joint text--image-to-point-cloud instance retrieval
    R@1, R@5, and R@10 (\%).}
    \label{fig:caption}
\end{figure}

\begin{figure}[!htbp]
    \centering
    \includegraphics[width=1\linewidth]{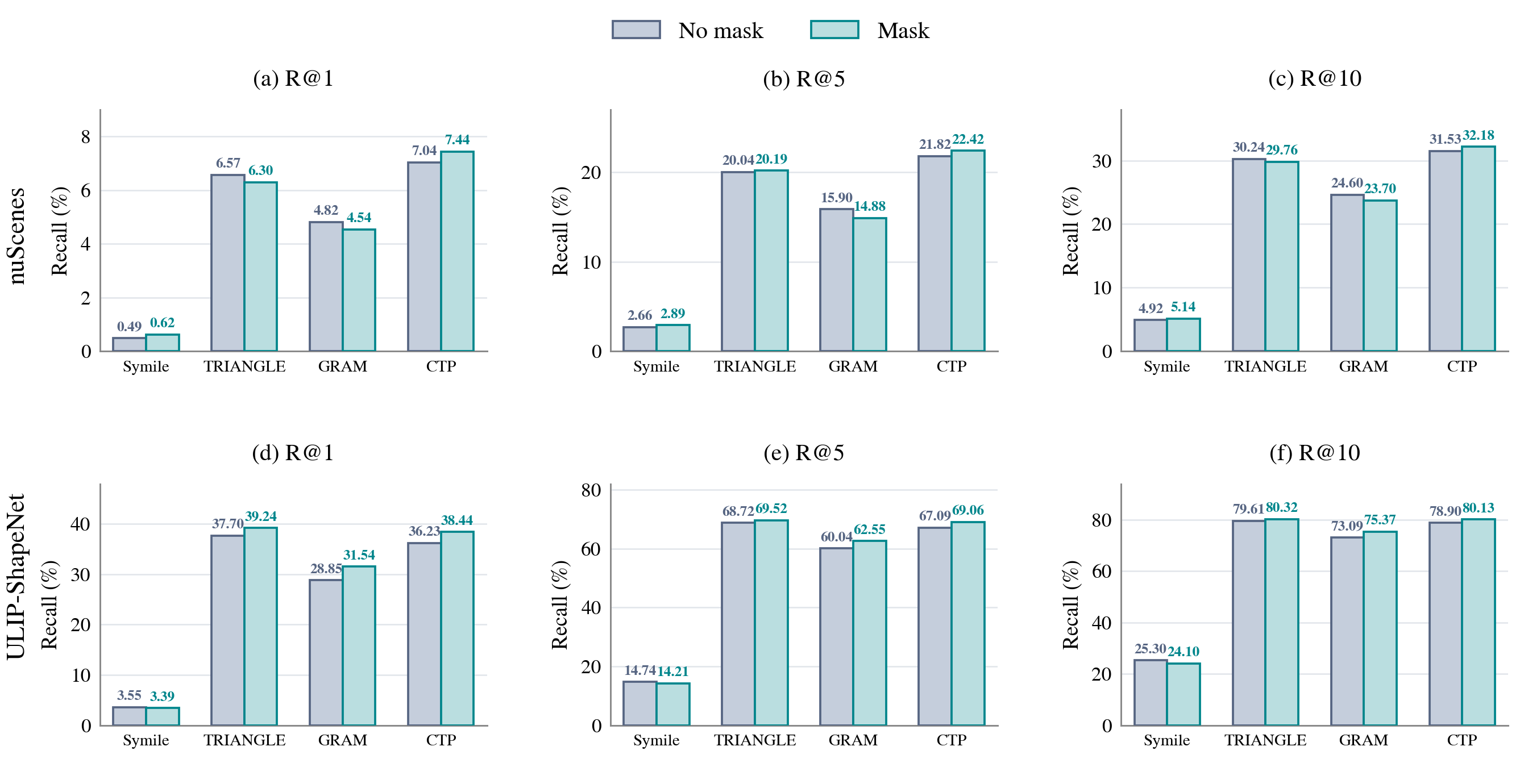}
    \caption{Effect of masking partial-match negatives on joint
    text--image-to-point-cloud instance retrieval
    R@1, R@5, and R@10 (\%) on nuScenes and ULIP-ShapeNet.}
    \label{fig:mask_ablation_appendix}
\end{figure}

\subsection{Additional Retrieval Results with Masking}
\label{sec:mask_retrieval}
Figure~\ref{fig:mask_ablation_appendix} complements the R@5 results in Figure~\ref{fig:mask_effect} by reporting R@1, R@5, and R@10.
For CTP, masking consistently improves all three metrics on both datasets.

\subsection{Comparison of Partial-Match Masking Strategies}
\label{sec:mask_strategy_ablation}

We compare three masking strategies for CTP: \textit{No mask} retains all tuple candidates, \textit{Row+Col} removes anchor-sharing partial-match negatives, and \textit{Row+Col+Diag} additionally removes off-anchor diagonal candidates containing a matched pair between the
two non-anchor modalities.
The fully matched positive tuple is retained in all settings.
All other training settings are kept identical.

Table~\ref{tab:mask_strategy_comparison} compares no masking, anchor-sharing masking (Row+Col), and masking all partial-match negatives (Row+Col+Diag).
Row+Col+Diag achieves slightly higher classification mAcc, while Row+Col consistently achieves the best retrieval results on both datasets and outperforms No mask across all reported
metrics.
This suggests that retaining off-anchor diagonal candidates provides useful instance-discrimination signals, motivating our adoption of anchor-sharing masking.

\begin{table*}[t]
\centering
\small
\setlength{\tabcolsep}{10pt}

\caption{
Effect of masking strategies on joint classification and text--image-to-point-cloud instance retrieval.
}
\label{tab:mask_strategy_comparison}

\vspace{3pt}

\begin{tabular}{lcccc}
\toprule
Mask strategy & mAcc & R@1 & R@5 & R@10 \\
\midrule
\multicolumn{5}{c}{nuScenes} \\
\midrule
No mask
    & 73.15 & \underline{7.04} & \underline{21.82} & \underline{31.53} \\
Row+Col+Diag
    & \textbf{75.38} & 5.65 & 18.81 & 28.14 \\
Row+Col
    & \underline{74.26} & \textbf{7.44}
    & \textbf{22.42} & \textbf{32.18} \\
\midrule
\multicolumn{5}{c}{ULIP-ShapeNet} \\
\midrule
No mask
    & 61.89 & 36.23 & 67.09 & 78.90 \\
Row+Col+Diag
    & \textbf{62.74} & \underline{37.97}
    & \underline{68.81} & \underline{79.00} \\
Row+Col
    & \underline{62.73} & \textbf{38.44}
    & \textbf{69.06} & \textbf{80.13} \\
\bottomrule
\end{tabular}

\end{table*}

\subsection{Effect of Point-Cloud Encoder Configurations}
\label{app:point_cloud_encoder}

\begin{table*}[t]
\centering
\small
\setlength{\tabcolsep}{9pt}

\caption{
Effect of point-cloud encoder configurations on joint classification and text--image-to-point-cloud instance retrieval.
}
\label{tab:backbone_comparison}

\vspace{3pt}

\begin{tabular}{llcccc}
\toprule
Backbone & Method & mAcc & R@1 & R@5 & R@10 \\
\midrule
\multicolumn{6}{c}{nuScenes} \\
\midrule

\multirow{2}{*}{LitePT-S~\citep{yue2026liteptlighterstrongerpoint}}
& ULIP-style~\citep{xue2023ulip}
    & 71.07 & 3.82 & 13.52 & 20.72 \\
& CTP (Ours)
    & \textbf{74.26} & \textbf{7.44}
    & \textbf{22.42} & \textbf{32.18} \\

\cmidrule(lr){1-6}

\multirow{2}{*}{PointNet++~\citep{qi2017pointnet++}}
& ULIP-style~\citep{xue2023ulip}
    & 67.21 & 3.24 & 11.62 & 18.07 \\
& CTP (Ours)
    & \textbf{72.81} & \textbf{4.71}
    & \textbf{15.53} & \textbf{23.03} \\

\midrule
\multicolumn{6}{c}{ULIP-ShapeNet} \\
\midrule

\multirow{2}{*}{LitePT-S~\citep{yue2026liteptlighterstrongerpoint}}
& ULIP-style~\citep{xue2023ulip}
    & 59.17 & 15.65 & 40.83 & 55.06 \\
& CTP (Ours)
    & \textbf{62.73} & \textbf{38.44}
    & \textbf{69.06} & \textbf{80.13} \\

\cmidrule(lr){1-6}

\multirow{2}{*}{PointNet++~\citep{qi2017pointnet++}}
& ULIP-style~\citep{xue2023ulip}
    & 61.66 & 22.75 & 52.98 & 66.90 \\
& CTP (Ours)
    & \textbf{64.07} & \textbf{42.48}
    & \textbf{72.29} & \textbf{81.93} \\

\bottomrule
\end{tabular}

\end{table*}

Table~\ref{tab:backbone_comparison} compares ULIP-style and CTP using LitePT-S and PointNet++ encoder configurations. CTP consistently outperforms ULIP-style across all metrics with both configurations, demonstrating that its gains extend beyond a single backbone. Under the evaluated configurations, LitePT-S performs better on sparse nuScenes point clouds, while PointNet++ achieves higher performance on dense ULIP-ShapeNet objects. We adopt LitePT-S for the main experiments given its stronger performance on sparse driving data.

\subsection{Bimodal Evaluation}
\label{sec:bimodal_evaluation}

To complement the joint evaluation, we assess point-cloud alignment with individual modalities.
Table~\ref{tab:text_point_classification} reports zero-shot point-cloud classification using class text prompts without image features.
CTP achieves the highest mAcc on nuScenes, KITTI, and Waymo. 
Table~\ref{tab:bimodal_retrieval} reports text-to-point-cloud and image-to-point-cloud instance retrieval.
ULIP-style performs best on text-to-point-cloud retrieval, while CTP achieves the highest image-to-point-cloud recall on both datasets, tying with TRIANGLE at R@10 on ULIP-ShapeNet.
These results demonstrate effective point-cloud alignment beyond joint inference, while highlighting differences between classification and instance retrieval.

\begin{table}[!htbp]
\centering
\caption{
Zero-shot point-cloud classification mAcc (\%) on nuScenes, KITTI, and Waymo.
Classification uses text--point-cloud cosine similarity without image features.
}
\label{tab:text_point_classification}

\vspace{3pt}

\small
\setlength{\tabcolsep}{7pt}

\begin{tabular}{lccc}
\toprule
Method & nuScenes & KITTI & Waymo \\
\midrule

ULIP-style~\citep{xue2023ulip}
    & 60.22 & 43.49 & 51.95 \\

Symile~\citep{saporta2024contrastingsymilesimplemodelagnostic}
    & 13.14 & 4.00 & 5.98 \\

TRIANGLE~\citep{cicchetti2025triangleenablesmultimodalalignment}
    & 57.16 & 32.08 & 36.99 \\

GRAM~\citep{cicchetti2025gramianmultimodalrepresentationlearning}
    & 52.66 & 29.47 & 46.48 \\

\midrule

CTP (Ours)
    & \textbf{68.39} & \textbf{50.44} & \textbf{63.11} \\

\bottomrule
\end{tabular}
\end{table}

\begin{table}[!htbp]
\centering
\caption{
Bimodal instance retrieval results (\%) on nuScenes and ULIP-ShapeNet.
T, I, and P denote instance captions, images, and point clouds, respectively.
$\dagger$: GRAM uses its two-modal Gram score,
$-\sqrt{1-\cos^2}$, while the other methods use cosine similarity.
}
\label{tab:bimodal_retrieval}

\vspace{3pt}

\small
\setlength{\tabcolsep}{7pt}

\begin{tabular}{lccc|ccc}
\toprule
\multirow{2}{*}{Method}
& \multicolumn{3}{c|}{T $\rightarrow$ P}
& \multicolumn{3}{c}{I $\rightarrow$ P} \\
\cmidrule(lr){2-4}\cmidrule(lr){5-7}
& R@1 & R@5 & R@10 & R@1 & R@5 & R@10 \\
\midrule
\multicolumn{7}{c}{nuScenes} \\
\midrule

ULIP-style~\citep{xue2023ulip}
& \textbf{1.11} & \textbf{4.27} & \textbf{7.62}
& 5.84 & 17.48 & 26.14 \\

Symile~\citep{saporta2024contrastingsymilesimplemodelagnostic}
& 0.02 & 0.09 & 0.23
& 0.11 & 0.62 & 1.12 \\

TRIANGLE~\citep{cicchetti2025triangleenablesmultimodalalignment}
& 0.49 & 2.37 & 4.28
& 6.78 & 20.34 & 30.12 \\

GRAM$^\dagger$~\citep{cicchetti2025gramianmultimodalrepresentationlearning}
& 0.09 & 0.28 & 0.53
& 2.83 & 10.26 & 16.40 \\
\midrule

CTP (Ours)
& 0.66 & 2.95 & 5.18
& \textbf{7.03} & \textbf{21.44} & \textbf{30.92} \\

\midrule
\multicolumn{7}{c}{ULIP-ShapeNet} \\
\midrule

ULIP-style~\citep{xue2023ulip}
& \textbf{1.12} & \textbf{4.84} & \textbf{8.41}
& 28.68 & 58.82 & 72.27 \\

Symile~\citep{saporta2024contrastingsymilesimplemodelagnostic}
& 0.02 & 0.24 & 0.43
& 0.02 & 0.15 & 0.20 \\

TRIANGLE~\citep{cicchetti2025triangleenablesmultimodalalignment}
& 0.68 & 3.41 & 5.88
& 37.86 & 68.90 & \textbf{79.77} \\

GRAM$^\dagger$~\citep{cicchetti2025gramianmultimodalrepresentationlearning}
& 0.03 & 0.17 & 0.41
& 19.37 & 47.12 & 60.76 \\
\midrule

CTP (Ours)
& 0.71 & 3.36 & 6.06
& \textbf{39.45} & \textbf{69.57} & \textbf{79.77} \\

\bottomrule
\end{tabular}
\end{table}

}

\subsection{Qualitative Results of Joint Retrieval}
\label{app:qualitative_retrieval}

\begin{figure}[!htbp]
    \centering
    \includegraphics[width=1\linewidth]{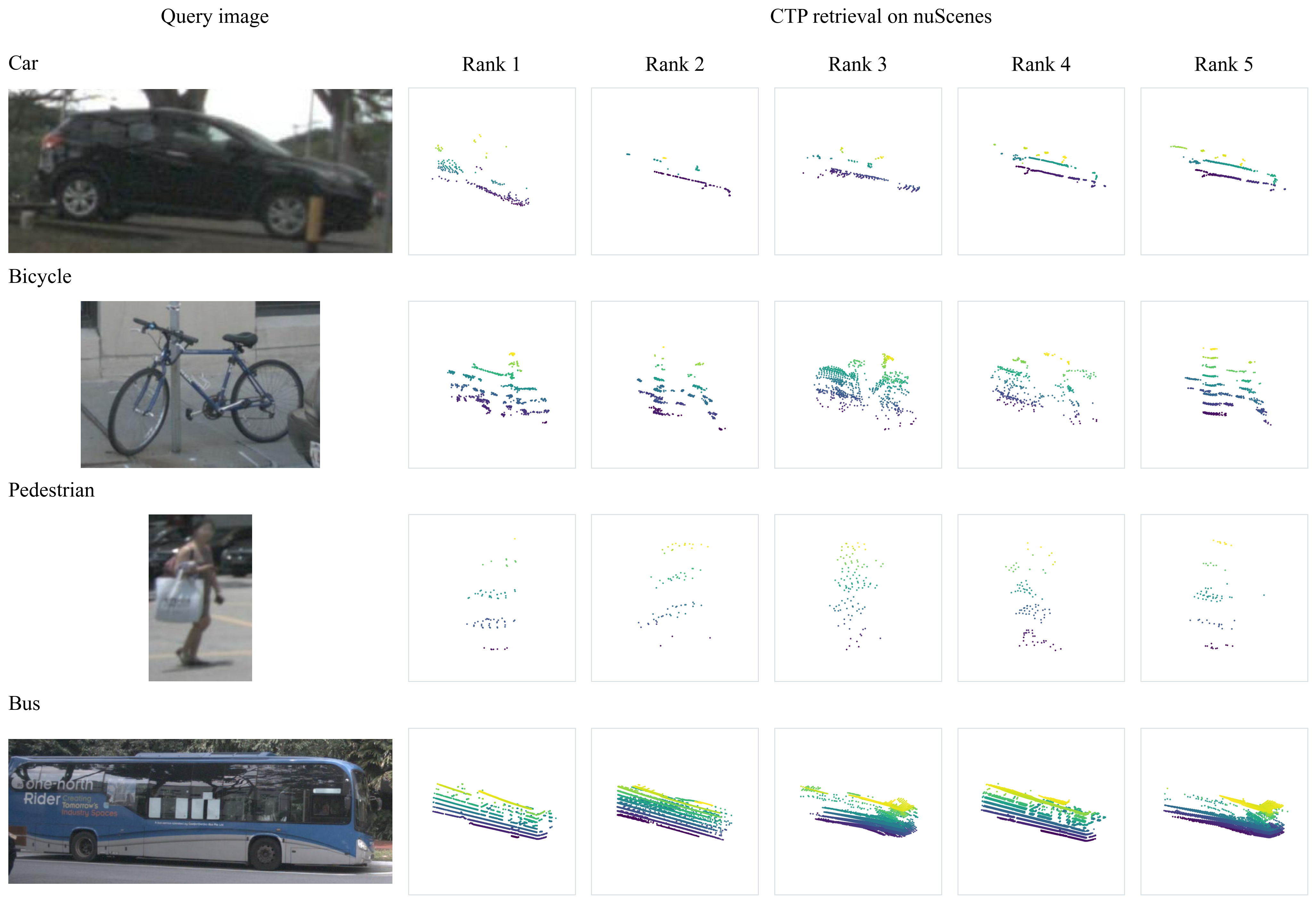}
    \caption{Qualitative examples of CTP joint text--image-to-point-cloud retrieval on nuScenes.
    Query captions are omitted for brevity.}
    \label{fig:nuscenes_joint_retrieval}
\end{figure}

\begin{figure}[!htbp]
    \centering
    \includegraphics[width=1\linewidth]{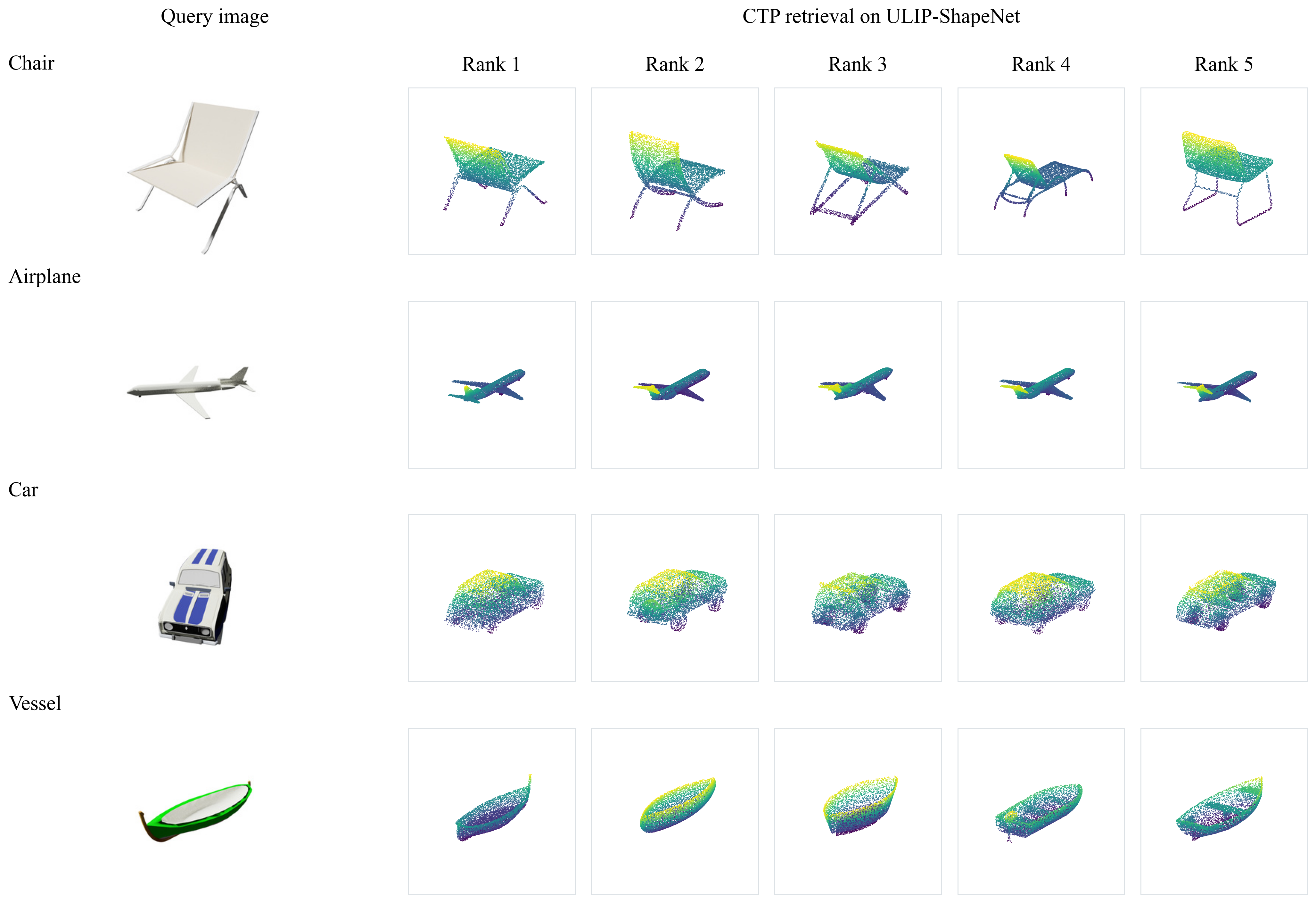}
    \caption{Qualitative examples of CTP joint text--image-to-point-cloud retrieval on ULIP-ShapeNet.
    Query captions are omitted for brevity.}
    \label{fig:shapenet_joint_retrieval}
\end{figure}

Figures~\ref{fig:nuscenes_joint_retrieval} and~\ref{fig:shapenet_joint_retrieval} show qualitative examples of CTP joint text--image-to-point-cloud retrieval on nuScenes and ULIP-ShapeNet. Each row presents a query image and the top five point clouds retrieved using the paired text and image as the joint query.

On ULIP-ShapeNet, the retrieved point clouds show visually similar object structures, such as the seat and backrest of chairs and the fuselage and wings of airplanes. On nuScenes, the examples illustrate retrieval with sparse and incomplete point clouds, where geometric details are less apparent. These examples show CTP's ability to retrieve relevant point clouds from joint text--image queries across both sparse and dense point clouds.

\end{document}